%% file: main.tex
\documentclass[10pt,twocolumn,twoside]{IEEEtran}
\IEEEoverridecommandlockouts
\usepackage{cite}
\usepackage{amsmath,amssymb,amsfonts}
\usepackage{graphicx}
\usepackage{textcomp}
\usepackage{xcolor}
\usepackage{bm}
\usepackage{cases}
\usepackage[skip=1pt, belowskip=0pt]{caption}
\usepackage[ruled,vlined,linesnumbered]{algorithm2e} 
    \usepackage{setspace} 
\usepackage[makeroom]{cancel} 
\usepackage{amsthm}
\usepackage{enumitem}

    \usepackage{physics}
    \usepackage{amsmath}
    \usepackage{tikz}
    \usepackage{mathdots}
    \usepackage{yhmath}
    \usepackage{cancel}
    \usepackage{color}
    \usepackage{siunitx}
    \usepackage{array}
    \usepackage{multirow}
    \usepackage{amssymb}
    \usepackage{gensymb}
    \usepackage{tabularx}
    \usepackage{booktabs}
    \usetikzlibrary{patterns}
    \usetikzlibrary{shapes}

\makeatletter
\def\BibTeX{{\rm B\kern-.05em{\sc i\kern-.025em b}\kern-.08em
    T\kern-.1667em\lower.7ex\hbox{E}\kern-.125emX}}
    
\allowdisplaybreaks
    
\include{macros}

\begin{document}
    
    \bstctlcite{IEEEexample:BSTcontrol}
    
    \title{
         Linear Regression with Distributed Learning: \\
         A Generalization Error Perspective
    }
    
    \author{
        \IEEEauthorblockN{
            Martin Hellkvist, 
            Ay\c ca \"Oz\c celikkale, 
            Anders Ahlén \\
            \thanks{%
                M. Hellkvist and A.~\"Oz\c celikkale acknowledges the support from Swedish Research Council under grant 2015-04011.
            }
            \thanks{
                This paper was presented in part at the 2021 ICML Workshop on Overparameterization: Pitfalls \& Opportunities, July, 2021.
            }
            \thanks{
                The authors are with the Department of Electrical Engineering at Uppsala University, Uppsala, Sweden
                (e-mail: \{Martin.Hellkvist, Ayca.Ozcelikkale, Anders.Ahlen\}@angstrom.uu.se).
            }
        }
    }
    
    \maketitle
    
    \newcommand\blfootnote[1]{%
        \begingroup
        \renewcommand\thefootnote{}\footnote{#1}%
        \addtocounter{footnote}{-1}%
        \endgroup
    }

    \input{abstract}
    
    \begin{IEEEkeywords}
    Distributed estimation, distributed optimization, supervised learning, generalization error, networked systems.  
    \end{IEEEkeywords}
    
    \input{my_cocoa}
    \input{definitions_theorems_etc}

    \input{intro}

    \input{notation}

    \setlength{\abovecaptionskip}{8pt} 
    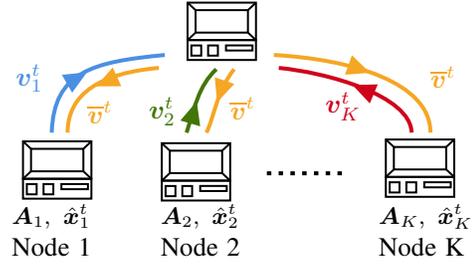
\begin{figure}[t]
        \centering
        \input{figures/system_fig_1}
        \caption{ 
            Distributed learning with \cocoa{}.
            }
        \kern-0.4em
        \label{fig:system_fig}
    \end{figure}
    \setlength{\abovecaptionskip}{2pt} 
    \input{problem_statement}
    \input{distributed_solution}
    \input{gen_error_main}

    \input{numerical_results}

\input{discussion}
    \input{conclusions}
    
    \input{appendix/appendix}

    \bibliographystyle{IEEEtran}
    \kern-0.5em
    \bibliography{ref}    
\end{document}

%% file: macros.tex
\newcommand{\Rbb}{\mathbb{R}}

\renewcommand{\Pr}{\mathbb{P}} 
\mathchardef\myhyphen="2D

\def\qed{\hfill $\square$}

\newcommand{\xhat}{\hat{\xvec}}
\newcommand{\xhatinit}{\hat{\xvec}^0}
\newcommand{\xhatt}{\hat{\xvec}^t}

\newcommand{\xhattp}{\hat{\xvec}^{t+1}}
\newcommand{\xhattm}{\hat{\xvec}^{t-1}}
\newcommand{\dx}{{\dxvec}}
\newcommand{\dxt}{{\dxvec}^t}
\newcommand{\xtilde}{\Tilde{\xvec}}
\newcommand{\xtildet}{\Tilde{\xvec}^t}

\newcommand{\cocoa}{\mbox{\textsc{CoCoA}}}

\newcommand{\sigmabar}{\bar{\sigma}}
\newcommand{\varphibar}{\bar{\varphi}}
\newcommand{\aggregationp}{\varphibar}
\newcommand{\subproblemp}{\sigmabar}
\newcommand{\egen}{\kappa}

\newcommand{\matr}[1]{\bm{#1}}
\newcommand{\Amat}{\matr{A}}
\newcommand{\Abar}{\bar{\matr{A}}}
\newcommand{\Bmat}{\matr{B}}
\newcommand{\Cmat}{\matr{C}}
\newcommand{\Dmat}{\matr{D}}

\newcommand{\Mmat}{\matr{M}}

\newcommand{\Rmat}{\matr{R}}
\newcommand{\Umat}{\matr{U}}
\newcommand{\Vmat}{\matr{V}}

\newcommand{\Zmat}{\matr{Z}}
\newcommand{\Sigmamat}{\matr{\Sigma}}
\newcommand{\Lambdamat}{\matr{\Lambda}}

\newcommand{\avec}{\matr{a}}
\newcommand{\cvec}{\matr{c}}
\newcommand{\dxvec}{\Delta\matr{x}}

\newcommand{\hvec}{\matr{h}}
\newcommand{\mvec}{\matr{m}}

\newcommand{\uvec}{\matr{u}}
\newcommand{\vvec}{\matr{v}}
\newcommand{\vbar}{\bar{\matr{v}}}
\newcommand{\wvec}{\matr{w}}
\newcommand{\xvec}{\matr{x}}
\newcommand{\yvec}{\matr{y}}
\newcommand{\zvec}{\matr{z}}

\newcommand{\omegavec}{\matr{\omega}}

\newcommand{\Imat}{\matr{I}}

\newcommand{\ith}{$i$\textsuperscript{th}}

\ifx \rank \undefined
    \newcommand{\rank}{\text{Rank}}
\else
    \renewcommand{\rank}{\text{Rank}}
\fi

\newcommand{\eye}[1]{\Imat_{#1}}

\newcommand{\Ebb}{\mathbb{E}}

\DeclareMathAlphabet{\mymathbb}{U}{BOONDOX-ds}{m}{n}

\newcommand{\T}{^\mathrm{T}}

\newcommand{\diag}{\text{diag}}

\newcommand{\p}{^+}

\newcommand{\inv}{^{-1}}

\ifx \norm \undefined
    \newcommand{\norm}[1]{\big\lVert#1\big\rVert}
\else
    \renewcommand{\norm}[1]{\big\lVert#1\big\rVert}
\fi

\let\oldin\in
\renewcommand{\in}{{\,\oldin\,}}

\let\oldnotin\notin
\renewcommand{\notin}{{\,\oldnotin\,}}

\renewcommand{\th}{\textsuperscript{th}}

\ifx \tr \undefined
    \newcommand{\tr}{\operatorname{tr}}
\else
    \renewcommand{\tr}{\operatorname{tr}}
\fi

\newcommand{\krange}{{k=1,\,\dots,\,K}}
\newcommand{\ikrange}{{i=1,\,\dots,\,K}}

\long\def\/*#1*/{}

\newcommand{\Kc }{\mathcal{K}}
\newcommand{\Nc }{\mathcal{N}}

\newcommand{\Sc }{\mathcal{S}}

\definecolor{myblue}{rgb}{0.22, 0.30, 0.70}
\definecolor{myblue2}{rgb}{0.0328, 0.0539, 0.4758}
\definecolor{mygreen2}{rgb}{ 0.030 0.400 0.050} 
\definecolor{mygreen3}{rgb}{ 0.0328 0.1758 0.0539} 
\definecolor{myred}{rgb}{0.4758, 0.0328, 0.0539}
\definecolor{myred2}{rgb}{0.75, 0.0328, 0.0539}

\newcommand{\rmaxi}{r_{\max,i}}
\newcommand{\rmini}{r_{\min,i}}
\newcommand{\rmaxk}{r_{\max,k}}
\newcommand{\rmink}{r_{\min,k}}
\newcommand{\sigmamin}{\sigma_{\min}}
\newcommand{\sigmamax}{\sigma_{\max}}
\newcommand{\lambdamin}{\lambda_{\min}}
\newcommand{\lambdamax}{\lambda_{\max}}

%% file: abstract.tex
\begin{abstract}
    Distributed learning provides an attractive framework for scaling the learning task by sharing the computational load over multiple nodes in a network.
    Here, we investigate the performance of distributed learning for large-scale linear regression where the model parameters, 
    i.e., the unknowns, are distributed over the network.
    We adopt a statistical learning approach.
    In contrast to works that focus on the performance on the training data, we focus on the generalization error, 
    i.e., the performance on unseen data. 
    We provide high-probability bounds on the generalization error for both isotropic and correlated Gaussian data as well as sub-gaussian data. 
    These results reveal the dependence of the 
    generalization performance on the partitioning of the model over the network. 
    In particular, our results show that the generalization error of the distributed solution can be substantially higher than that of the centralized solution even when the error on the training data is at the same level for both the centralized and distributed approaches. 
    Our numerical results illustrate the performance with both real-world image data as well as synthetic data. 
\end{abstract}%

%% file: my_cocoa.tex
\newcommand{\mycocoa}{
    \SetKw{KwEnd}{end}
	\textbf{Input}: Data matrix $\Amat$ distributed column-wise according to partitioning $\{p_1,\,\cdots,\,p_k\}$. 
    Observations $\yvec$. Regularization parameter $\lambda$, aggregation parameter $\!\aggregationp \in (0,1]$ and subproblem parameter $\subproblemp$.
            
    \textbf{Initialize:} $\xhatinit=0\in\Rbb^{p\times 1}$, $\vvec_k^0{=}0\in\Rbb^{p\times 1} \,\, \forall k$.
    
    \setstretch{1.05}
    \For{$t=0,\,1,\,\dots,\,T$}{
        $\vbar^t = \frac{1}{K}\sum_{k=1}^K\vvec_k^t$
        
        \For{$k \in \{1,\,2,\,\dots,\,K\}$}{
            $\cvec_k^t = \lambda\xhatt_k 
            - \Amat_k\T(\yvec - \vbar^t)$
            
            $\dxt_k = -(\subproblemp\Amat_k\T\Amat_k + \lambda\eye{p_k})\p\cvec_k^t$
            
            $\xhattp_k = \xhatt_k + \aggregationp\dxt_k$
            
            $\vvec_k^{t+1} = \vbar^t + \aggregationp K\Amat_k\dxt_k$
        }
    }
    \caption{Implementation of \cocoa{} \cite{smith_cocoa_nodate} for \eqref{eqn:problem}.%
    \label{alg:cocoa} 
    }
    }

%% file: definitions_theorems_etc.tex
\newtheorem{thm}{\bf{Theorem}}
\newtheorem{cor}{\bf{Corollary}}
\newtheorem{lem}{\bf{Lemma}}
\newtheorem{prop}{\bf{Proposition}}
\newtheorem{rem}{\bf{Remark}}

\theoremstyle{remark} 
\newtheorem{defn}{\bf{Definition}}[section]
\newtheorem{ex}{\bf{Example}}[section]
\newtheorem{myexp}{\bf{Experiment}}

\newenvironment{theorem}
{\par\noindent \thm \begin{itshape}\noindent}
{\end{itshape}}

\newenvironment{lemma}
{ \par\noindent  \lem \begin{itshape}\noindent}
{\end{itshape}
}

\newenvironment{corollary}
{\par\noindent  \cor \begin{itshape}\noindent}
{\end{itshape}
}

\newenvironment{remark}
{\par\noindent \rem \begin{itshape}\noindent}
{\end{itshape}
}

\newenvironment{definition}
{\par\noindent \defn \begin{itshape}\noindent}
{\end{itshape}}

\newenvironment{experiment}
{\vspace{3pt} \par\noindent \myexp \begin{itshape} \noindent}
{\end{itshape}\vspace{6pt}}

\newenvironment{example}
{\vspace{2pt} \par\noindent \ex} 
{\vspace{2pt}}

%% file: intro.tex
\section{Introduction}
\kern-0.25em
Distributed learning provides a framework for sharing the computational burden of large-scale learning tasks over multiple nodes while addressing growing concerns related to security and
data privacy 
    \cite{
    niknam_federated_2019,
    wang_privacy-preserving_2020}. 
Accordingly, the field of distributed learning is progressing rapidly due to the increasing need and interest from both industry and academia, with applications ranging from
    edge computing 
    \cite{chen_coded_2020, wang_adaptive_2019}
    to
    large-scale machine learning \cite{bottou2018optimization, bonawitz_towards_2019, NIPS2012_4687}.
In this article, we consider distributed learning from the point of view of generalization error and contribute to the field
by highlighting and characterizing potential pitfalls, and providing guidelines for best practice.

In particular, we consider the statistical learning problem where a set of training data $\{(y_i, \avec_i)\}_{i=1}^n$ from a certain distribution is used to train a model, 
i.e., estimate parameters in a specified model structure,
so that it correctly predicts the output $y_i \in \Rbb$ given the corresponding input $\avec_i \in\Rbb^{p\times 1}$.
The performance of the trained model is often measured by its \textit{training error}, 
i.e., the error that the model makes over the training data,
and more importantly its \textit{generalization error}, 
i.e., the error that the model makes when estimating $y$ using $\avec$ when a new pair $(y, \avec)$ comes from the same distribution as the training data.  
The generalization error is an inherent part of statistical learning frameworks,
where it is innately embedded in the expected error values,
and the expectation is taken with respect to the signal model. 
    The generalization error has been thoroughly studied for different centralized approaches,
    for instance in  minimum mean-square error estimation frameworks \cite{kay1993fundamentals}.
Recently, the dependence of the generalization error on the number of model parameters and the training sample size  has been investigated for a range of models, such as neural networks and decision trees,
and the ``double descent'' risk curve has been proposed \cite{ belkin_reconciling_2019}.
The generalization error associated with the least-squares estimate under isotropic Gaussian data and Fourier features with partial models \cite{breiman_how_nodate,belkin_two_2019},
the effect of regularization under data correlation \cite{nakkiran2020optimal, hastie2020surprises},
as well as sub-gaussian regressor distributions \cite{bartlett2020benign, Sahai_Harmless}
have been presented.
These works emphasize trade-offs between model complexity and training sample size in a centralized learning setting, particularly in overparametrized scenarios.

    The growing need for distributed learning has lead to the development of several methods, 
    e.g.,
    primal-dual methods \cite{Koppel_SaddlePoint_2015,bedi_koppel_ketan_asynchronous_2019}
    and dual decomposition methods \cite{Boyd_Distributed_Dual_Decomposition_2007, Terelius_Decentralized},
    where the alternating direction method of multipliers (ADMM) \cite{boyd2011distributed} stands out as one of the most extensively studied algorithms.
Accordingly, different aspects of distributed learning methods have been explored, 
including
privacy protecting methods \cite{wang_privacy-preserving_2020},
time-varying constraints \cite{paternain_distributed_2020},
adaptive network architectures \cite{zhao_distributed_2015},
and communication efficient methods such as the quantized stochastic gradient descent \cite{alistarh2017qsgd},
as well as novel metrics for communication efficiency \cite{magnusson_communication_2018}.
Trade-offs between computation and communication \cite{smith_cocoa_nodate} has been explored as well.
In the case of generalization error,
a significant part of the existing work for distributed learning is performed under the mean-square error estimation framework,
including Kalman filtering \cite{haotian_zhang_dynamic_2009, Khan2008Distributing}, 
least-mean squares \cite{LopesSayed2008DiffusionLMS, sayed2014adaptation}
and the affine projection algorithm \cite{LiChambers2009DistributedAPA}.
A characterization of the generalization error in the case of linear discriminant analysis is presented in \cite{varshney_generalization_2012}.
The average behaviour of the generalization error for regression is presented in \cite{HellkvistOzcelikkaleAhlen_distributed2020_spawc}
under isotropic Gaussian data.
    
Despite this vast interest in distributed learning, 
this line of work typically assumes that it is the sensor readings that are distributed over the network \cite{sayed2014adaptation},
in contrast to the scenario where the model unknowns are distributed over the network \cite{tsitsiklis_distributed_1984,Khan2008Distributing}. 
We address this gap by providing high probability bounds on the generalization error in a distributed linear regression problem under a broad family of training data distributions.
The setting with distributed unknowns is particularly suited to the problems with large number of unknowns 
\cite{smith_cocoa_nodate},
such as neural networks \cite{ben-nun_demystifying_2018}.
    Motivated by the recent results on overparameterization in linear regression \cite{nakkiran2020optimal,bartlett2020benign, hastie2020surprises, Sahai_Harmless},
    we pay special attention to the overparameterized setting where the number of unknowns governed by each node is larger than the number of observations.

    We consider the influential distributed learning algorithm \cocoa{}~\cite{smith_cocoa_nodate},
    developed from its predecessors \cocoa{}-v1~\cite{jaggi2014communication} and \cocoa{}$^+$~\cite{ma2017distributed}.
    \cocoa{} is applicable to convex optimization problems,
and allows the nodes to use any local solver of their choice for their local subproblems,
enabling the usage of solvers with variable accuracy and a flexible trade-off between computation and communication \cite{smith_cocoa_nodate}.
In \cite{smith_cocoa_nodate},
the convergence rate of \cocoa{} was quantified in terms of the convexity properties of the optimization problem and accuracy of local solvers.
In contrast to the work in \cite{smith_cocoa_nodate},
we focus on the generalization error of \cocoa{} and the effect of different data partitioning schemes over the nodes. 

In this article,  we show that the generalization error depends heavily on the partitioning of the model parameters among the nodes.
In particular, we have the following main contributions:
We provide bounds on the generalization error that hold with high probability for both isotropic Gaussian as well as correlated Gaussian data.
Furthermore, for block-correlated and underparameterized local problems with general covariance structure,
we generalize these results to sub-gaussian data,
which include the Bernoulli and uniform distributions as special cases.
For the isotropic Gaussian case, 
we compare these probabilistic results with analytical results on the average behaviour \cite{HellkvistOzcelikkaleAhlen_distributed2020_spawc},
    which we also extend to the setting with noisy training data in this article.
    We note that the presented results cover a wide set of distributions,
    compared to the scope of \cite{HellkvistOzcelikkaleAhlen_distributed2020_spawc},
    which is limited to the isotropic Gaussian distribution.
Our numerical results illustrate the generalization error performance  with both synthetic data from these distributions and real-world image data \cite{lecun-mnisthandwrittendigit-2010}. 

Our results highlight a typically overlooked relationship between the training and generalization error in distributed learning. %
These findings illustrate that distributed learning schemes can significantly amplify the gap between the training error and the generalization error. 
More precisely, 
a distributed solution with a training error that is on the same level as that of the centralized solution is not guaranteed to have a generalization error that is as low as that of the centralized solution.

The rest of the paper is organized as follows: 
Section~\ref{sec:problem}  and Section~\ref{sec:dist} present the problem formulation and the distributed solution approach, respectively. 
Section \ref{sec:gen_err} provides preliminary results on the generalization error.
In Section \ref{sec:gaussian:iid},
\ref{sec:gaussian:correlated} and \ref{sec:subgaussian},
we present the results for the isotropic Gaussian,
correlated Gaussian, and the sub-gaussian settings,
respectively.
The numerical results are presented in 
Section~\ref{sec:numerical}.
We present further discussions of our results in Section~\ref{sec:discussions} and conclude the article in  Section~\ref{sec:conclusions}.

%% file: notation.tex
\textbf{Notation:}
        We denote the Moore-Penrose pseudoinverse and the transpose of a matrix $\Amat$ as $\Amat\p$ and $\Amat\T$, respectively.
        The $p\times p$ identity matrix is denoted as $\eye{p}$.
        The positive semi-definite (p.s.d.) partial
        ordering for real symmetric matrices is denoted by  $\succeq$.
        We use $\|\cdot\|$ to denote either the spectral norm or the Euclidean norm, depending on whether the argument is matrix- or vector valued.
        Throughout the paper, 
        we often partition vectors by blocks of their entries, 
        and matrices either by their blocks of columns or rows. 
        For instance,
        the column-wise partitioning of a matrix
            $\Amat \in \Rbb^{n\times p}$ 
        into $K$ blocks
        is given by 
            $\Amat = [\Amat_1, \, \cdots, \, \Amat_K]$,
        with 
            $\Amat_k\in\Rbb^{n\times p_k}$.
        The row-wise partitioning of a vector  $\xvec \in \Rbb^{p \times 1}$  into $K$ blocks 
            $\xvec_k\in\Rbb^{p_k\times 1}$
        is given by 
            $\xvec=[\xvec_1;\,\cdots;\,\xvec_K]$,
            where the semicolon denotes row-wise separation.
         We use $\sigmamax(\cdot)$, $\sigmamin(\cdot)$ to denote the largest and smallest singular values of a matrix,
        and $\lambdamax(\cdot)$, $\lambdamin(\cdot)$ to denote the largest and smallest eigenvalues.
        The notation
            $(\cdot)_+$
        is used as a short-hand for $\max\{0,\cdot\}$.
        In expressions such as $(\cdot)_+^2$, 
        the $\max$ function takes precedence over the square,
        i.e., $(\cdot)_+^2 = ((\cdot)_+)^2$.

%% file: figures/system_fig_1.tex
\tikzset{every picture/.style={line width=0.75pt}} 

\begin{tikzpicture}[x=0.75pt,y=0.7pt,yscale=-1,xscale=1]

\draw   (130,4.71) -- (164.5,4.71) -- (164.5,24.96) -- (130,24.96) -- cycle ; \draw   (134.05,8.76) -- (160.45,8.76) -- (160.45,20.91) -- (134.05,20.91) -- cycle ; \draw   (130,4.71) -- (134.05,8.76) ; \draw   (164.5,4.71) -- (160.45,8.76) ; \draw   (164.5,24.96) -- (160.45,20.91) ; \draw   (130,24.96) -- (134.05,20.91) ;
\draw   (130,24.96) -- (164.5,24.96) -- (164.5,35.71) -- (130,35.71) -- cycle ;
\draw   (131.93,27.85) -- (136.89,27.85) -- (136.89,33.36) -- (131.93,33.36) -- cycle ;
\draw   (140.36,27.85) -- (145.31,27.85) -- (145.31,33.36) -- (140.36,33.36) -- cycle ;
\draw   (149.37,27.83) -- (162.75,27.83) -- (162.75,30.58) -- (149.37,30.58) -- cycle ;

\draw [line width=1.5]  [dash pattern={on 1.69pt off 2.76pt}]  (170,97) -- (210,97) ;
\draw [color={rgb, 255:red, 74; green, 144; blue, 226 }  ,draw opacity=1 ][line width=1.5]    (61.5,75) .. controls (62.5,41.71) and (82.5,36.71) .. (118.5,33) ;
\draw [shift={(79.27,42.38)}, rotate = 510.16] [fill={rgb, 255:red, 74; green, 144; blue, 226 }  ,fill opacity=1 ][line width=0.08]  [draw opacity=0] (11.61,-5.58) -- (0,0) -- (11.61,5.58) -- cycle    ;
\draw [color={rgb, 255:red, 245; green, 166; blue, 35 }  ,draw opacity=1 ][line width=1.5]    (115.5,40) .. controls (86.5,42.71) and (69.5,53.71) .. (69.5,75) ;
\draw [shift={(83.84,49.96)}, rotate = 335.62] [fill={rgb, 255:red, 245; green, 166; blue, 35 }  ,fill opacity=1 ][line width=0.08]  [draw opacity=0] (11.61,-5.58) -- (0,0) -- (11.61,5.58) -- cycle    ;
\draw [color={rgb, 255:red, 65; green, 117; blue, 5 }  ,draw opacity=1 ][line width=1.5]    (129.5,75) .. controls (133.5,61.71) and (135.5,49.71) .. (144.5,40.71) ;
\draw [shift={(134.63,58.08)}, rotate = 466.3] [fill={rgb, 255:red, 65; green, 117; blue, 5 }  ,fill opacity=1 ][line width=0.08]  [draw opacity=0] (11.61,-5.58) -- (0,0) -- (11.61,5.58) -- cycle    ;
\draw [color={rgb, 255:red, 245; green, 166; blue, 35 }  ,draw opacity=1 ][line width=1.5]    (153.5,40.71) .. controls (144.5,52.71) and (143.5,66.71) .. (139.5,75) ;
\draw [shift={(144.91,58.63)}, rotate = 288.83] [fill={rgb, 255:red, 245; green, 166; blue, 35 }  ,fill opacity=1 ][line width=0.08]  [draw opacity=0] (11.61,-5.58) -- (0,0) -- (11.61,5.58) -- cycle    ;
\draw [color={rgb, 255:red, 208; green, 2; blue, 27 }  ,draw opacity=1 ][line width=1.5]    (243,75) .. controls (242.5,54.71) and (199.5,42.33) .. (176,40) ;
\draw [shift={(213.2,49.27)}, rotate = 382.6] [fill={rgb, 255:red, 208; green, 2; blue, 27 }  ,fill opacity=1 ][line width=0.08]  [draw opacity=0] (11.61,-5.58) -- (0,0) -- (11.61,5.58) -- cycle    ;
\draw [color={rgb, 255:red, 245; green, 166; blue, 35 }  ,draw opacity=1 ][line width=1.5]    (173.5,33) .. controls (212.5,35.33) and (252.5,47.71) .. (252.5,75) ;
\draw [shift={(222.07,42.62)}, rotate = 199.76] [fill={rgb, 255:red, 245; green, 166; blue, 35 }  ,fill opacity=1 ][line width=0.08]  [draw opacity=0] (11.61,-5.58) -- (0,0) -- (11.61,5.58) -- cycle    ;
\draw   (47,79.71) -- (81.5,79.71) -- (81.5,99.96) -- (47,99.96) -- cycle ; \draw   (51.05,83.76) -- (77.45,83.76) -- (77.45,95.91) -- (51.05,95.91) -- cycle ; \draw   (47,79.71) -- (51.05,83.76) ; \draw   (81.5,79.71) -- (77.45,83.76) ; \draw   (81.5,99.96) -- (77.45,95.91) ; \draw   (47,99.96) -- (51.05,95.91) ;
\draw   (47,99.96) -- (81.5,99.96) -- (81.5,110.71) -- (47,110.71) -- cycle ;
\draw   (48.93,102.85) -- (53.89,102.85) -- (53.89,108.36) -- (48.93,108.36) -- cycle ;
\draw   (57.36,102.85) -- (62.31,102.85) -- (62.31,108.36) -- (57.36,108.36) -- cycle ;
\draw   (66.37,102.83) -- (79.75,102.83) -- (79.75,105.58) -- (66.37,105.58) -- cycle ;

\draw   (117,80.71) -- (151.5,80.71) -- (151.5,100.96) -- (117,100.96) -- cycle ; \draw   (121.05,84.76) -- (147.45,84.76) -- (147.45,96.91) -- (121.05,96.91) -- cycle ; \draw   (117,80.71) -- (121.05,84.76) ; \draw   (151.5,80.71) -- (147.45,84.76) ; \draw   (151.5,100.96) -- (147.45,96.91) ; \draw   (117,100.96) -- (121.05,96.91) ;
\draw   (117,100.96) -- (151.5,100.96) -- (151.5,111.71) -- (117,111.71) -- cycle ;
\draw   (118.93,103.85) -- (123.89,103.85) -- (123.89,109.36) -- (118.93,109.36) -- cycle ;
\draw   (127.36,103.85) -- (132.31,103.85) -- (132.31,109.36) -- (127.36,109.36) -- cycle ;
\draw   (136.37,103.83) -- (149.75,103.83) -- (149.75,106.58) -- (136.37,106.58) -- cycle ;

\draw   (230,78.71) -- (264.5,78.71) -- (264.5,98.96) -- (230,98.96) -- cycle ; \draw   (234.05,82.76) -- (260.45,82.76) -- (260.45,94.91) -- (234.05,94.91) -- cycle ; \draw   (230,78.71) -- (234.05,82.76) ; \draw   (264.5,78.71) -- (260.45,82.76) ; \draw   (264.5,98.96) -- (260.45,94.91) ; \draw   (230,98.96) -- (234.05,94.91) ;
\draw   (230,98.96) -- (264.5,98.96) -- (264.5,109.71) -- (230,109.71) -- cycle ;
\draw   (231.93,101.85) -- (236.89,101.85) -- (236.89,107.36) -- (231.93,107.36) -- cycle ;
\draw   (240.36,101.85) -- (245.31,101.85) -- (245.31,107.36) -- (240.36,107.36) -- cycle ;
\draw   (249.37,101.83) -- (262.75,101.83) -- (262.75,104.58) -- (249.37,104.58) -- cycle ;

\draw (40,112) node [anchor=north west][inner sep=0.75pt]  [font=\small]  {$\boldsymbol{A}_{1} ,\ \hat{\boldsymbol{x}}^{t}_{1}$};
\draw (42,36.4) node [anchor=north west][inner sep=0.75pt]  [color={rgb, 255:red, 74; green, 144; blue, 226 }  ,opacity=1 ]  {$\boldsymbol{v}^{t}_{1}$};
\draw (115,112) node [anchor=north west][inner sep=0.75pt]  [font=\small]  {$\boldsymbol{A}_{2} ,\ \hat{\boldsymbol{x}}^{t}_{2}$};
\draw (225,112) node [anchor=north west][inner sep=0.75pt]  [font=\small]  {$\boldsymbol{A}_{K} ,\ \hat{\boldsymbol{x}}^{t}_{K}$};
\draw (78,56.4) node [anchor=north west][inner sep=0.75pt]  [color={rgb, 255:red, 245; green, 166; blue, 35 }  ,opacity=1 ]  {$\overline{\boldsymbol{v}}^{t}$};
\draw (109,54.4) node [anchor=north west][inner sep=0.75pt]  [color={rgb, 255:red, 65; green, 117; blue, 5 }  ,opacity=1 ]  {$\boldsymbol{v}^{t}_{2}$};
\draw (150,54.4) node [anchor=north west][inner sep=0.75pt]  [color={rgb, 255:red, 245; green, 166; blue, 35 }  ,opacity=1 ]  {$\overline{\boldsymbol{v}}^{t}$};
\draw (198,51.4) node [anchor=north west][inner sep=0.75pt]  [color={rgb, 255:red, 208; green, 2; blue, 27 }  ,opacity=1 ]  {$\boldsymbol{v}^{t}_{K}$};
\draw (251,37.4) node [anchor=north west][inner sep=0.75pt]  [color={rgb, 255:red, 245; green, 166; blue, 35 }  ,opacity=1 ]  {$\overline{\boldsymbol{v}}^{t}$};
\draw (40,130) node [anchor=north west][inner sep=0.75pt]   [align=left] {Node 1};
\draw (115,130) node [anchor=north west][inner sep=0.75pt]   [align=left] {Node 2};
\draw (225,130) node [anchor=north west][inner sep=0.75pt]   [align=left] {Node K};

\end{tikzpicture}

%% file: problem_statement.tex
\section{Problem Statement}\label{sec:problem}
We focus on the linear model
\begin{equation}\label{eqn:model}
    y_i = \avec_i\T \xvec + w_i,
\end{equation}
where $y_i\in \Rbb$ is the \ith{} observation,
$\avec_i\in \Rbb^{p\times 1}$ is the \ith{} regressor,
$w_i\in \Rbb$ is the corresponding unknown disturbance,
and $\xvec\in \Rbb^{p\times 1}$ is the vector of unknown model parameters.
We consider the problem of estimating $\xvec$ given $n$ pairs of observations and regressors, 
i.e., the training dataset
$ \{(y_i, \avec_i)\}_{i=1}^n$ 
by minimizing the following regularized cost function: 
    \begin{equation}\label{eqn:problem}
        \min_{\xvec\in\Rbb^{p\times1}} \frac{1}{2}\norm{\yvec - \Amat\xvec}^2 + \frac{\lambda}{2} \norm{\xvec}^2,
    \end{equation}
where $\Amat\in\Rbb^{n\times p}$ is the regressor matrix whose $i$\textsuperscript{th} row is given by $\avec_i\T\in\Rbb^{1\times p}$,
$\yvec\in\Rbb^{n\times 1}$ is the vector of observations $y_i$,
and $\lambda \geq 0$ is a regularization parameter.

We consider the setting where the regressors $\avec_i\T$ are independent and identically distributed (i.i.d.) zero-mean random vectors with a given distribution $\mathcal{D}(\Sigmamat)$,
with the covariance matrix 
$\Sigmamat = \Ebb_{\avec_i}[ \avec_i \avec_i\T ]\in\Rbb^{p\times p}$.
Under this regressor model, 
we investigate the generalization error of the solution to \eqref{eqn:problem} found by the distributed solver \cocoa{} \cite{smith_cocoa_nodate}.
    In order to simplify the theoretical analysis, we mainly consider the unregularized and noise-free setting,
    i.e., with $\lambda=0$ and $\wvec=0$. 
    Under these simplifications,
    we derive bounds illustrating how the generalization error  depends on the partitioning of the model over the network. 
    In order to provide background for these results,
    we consider the more general case with $\wvec\neq 0$ in Sections~\ref{sec:problem}~--~\ref{sec:gen_err} together with a discussion on the case with $\lambda>0$.
    In Section~\ref{sec:gen_err} and Section~\ref{sec:gauss:iid:expected}, we provide insights about why the training noise does not necessarily weaken the dependence of generalization error on partitioning.
    For the regularized case, 
    i.e., with $\lambda>0$,  
    and for the case with non-zero noise, 
    we provide numerical results which illustrate how the same heavy dependence on partitioning occurs for $\lambda>0$ before convergence;  
    and for $\wvec \neq 0$ even after convergence, see Section~\ref{sec:numerical}.
In the remainder of this section, we define the generalization error.
We provide details for \cocoa{} in Section~\ref{sec:dist}.

    Let $\xhat$ be an estimate of $\xvec$ found using a given set of training data $ \{(y_i, \avec_i)\}_{i=1}^n$,
    where $\avec_i \sim \mathcal{D}(\Sigmamat)$ and $y_i=\avec_i\T\xvec+w_i$.
    Let $(y,\avec)$ be a new input-output pair with 
    $\avec \sim \mathcal{D}(\Sigmamat)$ and $y =\avec\T\xvec + w$.
    Then,
    the generalization error is given by 
        \begin{align}
            \egen(\xhat) 
            = & \Ebb_{\avec}[(\avec\T\xvec - \avec\T\xhat)^2 ]\\
            = & (\xvec-\xhat)\T \Ebb_{\avec}[\avec \avec\T](\xvec-\xhat)
                \label{eqn:test:crossvanish} \\
            = &(\xvec-\xhat)\T \Sigmamat (\xvec-\xhat),
            \label{eqn:gen_err}
        \end{align}
    where we have used that $\xhat$ is a fixed estimate under the given training data.  
    The notation $\Ebb_{\avec}[\cdot]$ is used to emphasize that the expectation is over the previously unseen regressor $\avec$.
   One may alternatively consider the prediction error in $y$ instead of $\avec\T\xvec$:
    \begin{equation}\label{eqn:gen_err_noisy_def}
        \Ebb_{\avec,w}[(y-\avec\T\xhat)^2] = \egen(\xhat) + \sigma_w^2,
    \end{equation}
    where the noise $w$ in the test data is assumed to be zero-mean with variance $\sigma_w^2$ and statistically independent with the regressor $\avec$. 
    Since the noise in test data just gives an additive, irreducible term,
    we focus directly on $\egen(\xhat)$ in our technical development.
We are interested in the behaviour of the generalization error  $\egen(\xhat) \in\Rbb$ with respect to the distribution of the training data, 
i.e., $\Amat$, 
and the partitioning of the data over the nodes.

In the centralized case,
a solution to \eqref{eqn:problem} is found as 
\setlength{\abovedisplayskip}{2pt}
\setlength{\belowdisplayskip}{2pt} 
\begin{equation}\label{eqn:centralized_sol}
    \xhat_C = (\Amat\T\Amat + \lambda\eye{p})\p \Amat\T \yvec.
\end{equation}%
\setlength{\abovedisplayskip}{4pt}%
\setlength{\belowdisplayskip}{4pt}%
In general, with $\lambda=0$, 
there can exist multiple solutions to \eqref{eqn:problem}.
With the Moore-Penrose pseudoinverse,
the solution with the minimum Euclidean norm is obtained. 

%% file: distributed_solution.tex
\section{Distributed Solution Approach}\label{sec:dist}
        \setlength{\textfloatsep}{0pt}
        \begin{algorithm}[t]
            \mycocoa
        \end{algorithm}
    We now discuss how to obtain a solution  $\xhat$ for \eqref{eqn:problem} using the distributed solution approach  \cocoa{}  \cite{smith_cocoa_nodate}, see Figure~\ref{fig:system_fig} and
    Algorithm~\ref{alg:cocoa}.
    Here,
    mutually exclusive subsets of the $p$ unknown parameters in $\xvec$ and the associated subset of columns in $\Amat\in\Rbb^{n\times p}$ are distributed over $K$ nodes, $K \leq p$. 
    Hence, 
    node $k$ governs the learning of $p_k$ variables,
    denoted by $\xvec_k\in\Rbb^{p_k\times 1}$, 
    where $\sum_{k=1}^K p_k = p$.
    We denote the part of $\Amat$ available at node $k$ as $\Amat_k\in\Rbb^{n\times p_k}$.
    All nodes have access to the vector of observations $\yvec\in\Rbb^{n\times 1}$.
        Using this partitioning,
        $\yvec = \Amat \xvec + \wvec$ can be expressed as 
        \setlength{\abovedisplayskip}{1pt}
        \setlength{\belowdisplayskip}{1pt}  
        \begin{align}
         \yvec 
            = [\Amat_1,\cdots,\Amat_K]
            \begin{bmatrix}
                \xvec_1\\\vdots\\\xvec_K
            \end{bmatrix} + \wvec
            = \sum_{k=1}^K \Amat_k \xvec_k + \wvec,
        \end{align}
        \setlength{\abovedisplayskip}{4pt}%
        \setlength{\belowdisplayskip}{4pt}%
    Note that the submatrices $\Amat_k$'s and the observation vector $\yvec$ are fixed over all iterations. 
    
    Node $k$ forms an estimate of $\xvec_k$ using $\yvec$, $\Amat_k$ and a centrally computed variable  $\vbar^t\in\Rbb^{n\times 1}$. 
    Let $\xhatt_k\in\Rbb^{p_k\times 1}$  denote the estimate of $\xvec_k$ at node $k$
    and iteration $t$. 
    Accordingly, let $\xhatt=[\xhatt_1;\ldots; \xhatt_k] \in\Rbb^{p\times 1}$ denote the estimate of $\xvec$ at iteration $t$.
    At iteration $t$, 
    node $k$ receives the centrally computed variable $\vbar^t$,
    which it uses to compute $\dx^t_k\in\Rbb^{p_k\times 1}$ (Line 6-7, Alg.~\ref{alg:cocoa}),
    i.e., the update for $\xhatt_k$ (Line 8).
    The node keeps track of its contribution for estimating $\yvec$
    by computing the local estimate $\vvec_k^{t+1}\in\Rbb^{n\times 1}$,
    using 
    $\vbar^t$ and $\dx_k^t$ (Line 9).
    Then, the variable $\vvec_k^{t+1}$ is sent to a central node to 
    create $\vbar^{t+1}$ (Line 4).
    The central node then sends $\vbar^{t+1}$ to the nodes and the next iteration begins.
    
    We now explain how node $k$ finds the update $\dx_k^t$.
    To find $\dx_k^t$,
    \cocoa{} solves the following convex minimization problem at each node \cite{smith_cocoa_nodate}:

        \setlength{\abovedisplayskip}{2pt}
        \setlength{\belowdisplayskip}{2pt}    
        \setlength{\jot}{-5pt}
        \begin{align}\label{eqn:problem_t}
        \begin{split}
            \min_{\dx^t_k} 
            \tfrac{1}{K} f(\vbar^t)
            & + \nabla_{\vbar^t} f(\vbar^t)\T
                \Amat_k\dx^t_k \\
            & + \tfrac{\subproblemp}{2\tau}
                \norm{\Amat_k\dx^t_k}^2
            +\tfrac{\lambda}{2}
                \norm{\xhatt_k+\dx^t_k}^2,
        \end{split}
        \end{align}
        \setlength{\jot}{0pt}%
        \setlength{\abovedisplayskip}{4pt}%
        \setlength{\belowdisplayskip}{4pt}%
        where 
        $f(\vbar^t)=\tfrac{1}{2}\norm{\yvec - \vbar^t}^2$
        is the first term of the objective function in \eqref{eqn:problem},  evaluated at $\vbar^t$.
        Note that $\vbar^t =\Amat\xhatt$ by Algorithm~\ref{alg:cocoa}.
        The first two terms of  \eqref{eqn:problem_t} comes from the linearization of $f(\cdot)$ around the current value of $\vbar^t$,
        and the third term  $\frac{\subproblemp}{2\tau}\|\Amat\dxt_k\|^2$ penalizes large changes in $\vbar^t =\Amat\xhatt$.  
        The last term  
        $\frac{\lambda}{2}\|\xhatt_k+\dxt_k\|^2$
        corresponds to the local component of the regularization term in \eqref{eqn:problem} evaluated at $\xhatt_k+\dxt_k$.  
    
    The smoothness parameter for $f(\cdot)$ is  $\tau=1$\cite{smith_cocoa_nodate}.
    Only keeping the terms that depend on $\dx^t_k$ reveals that
    \eqref{eqn:problem_t} can be equivalently 
    solved by
        \setlength{\jot}{-5pt}
        \begin{align}\begin{split}\label{eqn:problem_open}
            \min_{\dx^t_k} & ~ (\dx^t_k)\T(
                \tfrac{\subproblemp}{2}\Amat_k\T\Amat_k 
                + \tfrac{\lambda}{2} \eye{p_k}
            )\dx^t_k\\
            & + ( \lambda\xhatt_k - \Amat_k\T(\yvec - \vbar^t) )\T\dx^t_k.
        \end{split}
        \end{align}
        \setlength{\jot}{1pt}%
    Taking the derivative with respect to $\dx^t_k$ and setting it to zero, we obtain
        \begin{align}\begin{split}\label{eqn:iteration_eqn_system}
            (\subproblemp & \Amat_k\T\Amat_k + \lambda \eye{p_k})\dx^t_k = - ( \lambda\xhatt_k - \Amat_k\T(\yvec - \vbar^t) ).
        \end{split}\end{align}
    With $\lambda=0$,
    the existence of a matrix inverse is not guaranteed,
    in general.   
    Hence, the local solvers use Moore-Penrose pseudoinverse  to  solve  \eqref{eqn:iteration_eqn_system} to obtain 
        \begin{equation}\label{eqn:pinv_solver}
            \dx^t_k = - (\subproblemp\Amat_k\T\Amat_k + \lambda \eye{p_k})\p ( \lambda\xhatt_k - \Amat_k\T(\yvec - \vbar^t) ).
        \end{equation}
    The resulting algorithm for estimating $\xvec$ iteratively is presented in Algorithm~\ref{alg:cocoa}. 
    
    Similar to dual decomposition methods \cite{Boyd_Distributed_Dual_Decomposition_2007, Terelius_Decentralized} and in particular ADMM \cite{boyd2011distributed},    \cocoa{} encourages a consensus over nodes by utilizing Lagrangian duality.  
    Although it  is inherently connected to ADMM \cite{boyd2011distributed},  \cocoa{} utilizes a more simple update for $\hat{x}$ and allows  approximate proximal steps, see for instance  \cite[Eqn. (12), Eqn. (15)]{smith_cocoa_nodate}.

    \setlength{\abovedisplayskip}{4pt}%
    \setlength{\belowdisplayskip}{4pt}%

%% file: gen_error_main.tex
\section{Generalization Error with \cocoa{}}\label{sec:gen_err}
We are interested in  the behaviour of the generalization error in \eqref{eqn:gen_err} with respect to different partitioning schemes $\{p_1,\cdots,p_K\}$, as well as different distributions of the data.
For the rest of the article, 
we consider the case with $\lambda=0$, 
except for the numerical experiments in Section~\ref{sec:numerical}.
We set $\subproblemp=\aggregationp K$, 
as it is considered a safe choice in terms of convergence\cite[Sec.~3.1]{smith_cocoa_nodate},  
and provide additional experiments with other values in Section~\ref{sec:numerical_hyperparam}. 
As shown in Lemma~1 of \cite{HellkvistOzcelikkaleAhlen_distributed2020_spawc},
the iterations of Algorithm~\ref{alg:cocoa} can be expressed as
\begin{align}\label{eqn:xhattp:x}
    \xhattp = \Bmat \xhatt + \tfrac{1}{K} \Abar \yvec,
\end{align}
where $ \Bmat \in\Rbb^{p\times p}$ and $\Abar\in\Rbb^{p\times n}$ are given by
\begin{align}
    \label{eqn:xhattp:B}
    \Bmat &= \left(\eye{p} - \tfrac{1}{K} \Abar \Amat\right),
\end{align}
and
\begin{align}
    \Abar &= 
        \begin{bmatrix} 
            \Amat_1\p\\ \Amat_2\p \\ \vdots\\ \Amat_K\p  
        \end{bmatrix}. \label{eqn:xhattp:Abar}
\end{align}
Note that, under $\subproblemp=\aggregationp K$, $\aggregationp$ and $\subproblemp$ enters into \eqref{eqn:xhattp:x} as $\frac{\aggregationp}{\subproblemp}=\frac{1}{K}$, hence the solutions $\xhatt$ are independent of the particular values of 
$\aggregationp$ and $\subproblemp$.

Using \eqref{eqn:xhattp:x}-\eqref{eqn:xhattp:B} and $\yvec=\Amat\xvec+\wvec$,
the error vector $\xtildet =  \xvec - \xhatt $ can be expressed as 
\begin{align}
    \xtildet 
    & = \xvec - \Bmat\xhattm - \tfrac{1}{K}\Abar \yvec 
     = \Bmat \xtilde^{t-1} - \tfrac{1}{K}\Abar \wvec\\
    & = \Bmat^2 \xtilde^{t-2} - (\Bmat+\eye{p})\tfrac{1}{K}\Abar \wvec = \cdots\\
    & = \Bmat^t \xvec - \tfrac{1}{K}\sum_{i=0}^{t-1} \Bmat^i\Abar \wvec
    \triangleq \Bmat^t \xvec - \Rmat_{t} \wvec,
    \label{eqn:gen_err_B}%
\end{align}
where we have defined $ \Rmat_{t} = \tfrac{1}{K}\sum_{i=0}^{t-1} \Bmat^i\Abar$, and used the fact that the algorithm is initialized using $\xhat^{0}=0$,
thus $\xtilde^0 = \xvec - \xhat^{0} = \xvec$.
Hence, the evolution of the error vector is governed by the matrix $\Bmat$. 
Using \eqref{eqn:gen_err} and \eqref{eqn:gen_err_B},
we can bound the generalization error $\egen(\xhatt)$ as
\setlength{\abovedisplayskip}{1pt}
\setlength{\belowdisplayskip}{1pt}    
\begin{align}\allowdisplaybreaks
    \egen(\xhatt) 
    & = \| \Sigmamat^{1/2} (\Bmat^t\xvec - \Rmat_t\wvec)\|^2 
        \label{eqn:noisy_egen_Bbound:termstogether} \\
    & \leq 
        2\|\Sigmamat\| 
        (\|\Bmat\|^{2t} \|\xvec\|^2 
        + \|\Rmat_t\|^{2} \|\wvec\|^2)
        ,\label{eqn:noisy_egen_Bbound}
\end{align}
\setlength{\abovedisplayskip}{4pt}%
\setlength{\belowdisplayskip}{4pt}%
where $\|\cdot\|$ denotes the spectral norm for matrices, and the Euclidean norm for vectors. Here, we have used properties of the  matrix/vector norms and the fact that $ \| \Sigmamat^{1/2} \|^2 = \| \Sigmamat \|$ for $\Sigmamat \succeq 0$.

Considering the noise-free setting,
i.e., $\wvec=0$,
we obtain the bound
\setlength{\abovedisplayskip}{1pt}
\setlength{\belowdisplayskip}{1pt}    
\begin{align}\allowdisplaybreaks
    \egen(\xhatt) 
    = \| \Sigmamat^{1/2}   \xtildet\|^2 
    & = \| \Sigmamat^{1/2} \Bmat^t\xvec\|^2 
    \label{eqn:egen_iter_t}\\
    & \leq 
        \|\Sigmamat\| 
        \|\Bmat\|^{2t} \|\xvec\|^2.
        \label{eqn:egen_Bbound}
\end{align}
\setlength{\abovedisplayskip}{4pt}%
\setlength{\belowdisplayskip}{4pt}%
In the upcoming sections, 
we investigate the generalization error in terms of  the behaviour of $\| \Bmat \|$ for different statistical models for $\Amat$. 
As our results in the coming sections illustrate,
the spectral properties of $\Bmat=\eye{p} - \frac{1}{K}\Abar \Amat$
can heavily depend on the partitioning parameters $p_k$,
$\krange$.
In particular, if any $p_k$ is close to $n$, then $\|\Bmat\|$ cannot be bounded with high probability,
which is directly
reflected in the generalization error $\egen(\xhatt)$.
Both the noisy and the noise-free setting are  studied in the numerical results of Section~\ref{sec:numerical}.
The presented results illustrate how 
the error can be extremely large also with noisy training data,
hence the insights gained from our analytical study of the bound in \eqref{eqn:egen_Bbound} are also relevant for the noisy setting of \eqref{eqn:noisy_egen_Bbound:termstogether}.

We now compare generalization error and the training error. For $\wvec=0$, 
the training error associated with $\xhatt$, i.e.,
the error in reconstructing $\{ \avec_i\T \xvec\}_{i=1}^n$,  can be expressed as follows 
\begin{align}\label{eqn:training_error}
    \frac{1}{n} \sum_{i=1}^n (\avec_i\T \xvec - \avec_i\T \xhatt)^2
    & = \frac{1}{n} \| \Amat \xtildet \|^2 
    =  \frac{1}{n} \|\Amat \Bmat^t \xvec \|^2.
\end{align}
    Note that for the training error in \eqref{eqn:training_error},
    $\xtildet = \Bmat^t \xvec $ is multiplied with the  current realization $\Amat$. 
    On the other hand, 
    for the generalization error in 
    \eqref{eqn:noisy_egen_Bbound:termstogether}, there is a multiplication with $\Sigmamat^{1/2}$ due to averaging over realizations of the regressor matrix. 
This distinction can lead to a significant gap between the training and generalization error. 
We illustrate later in this section,
see \eqref{eqn:training_error_broad_case},
that the training error is exactly zero under certain partitioning schemes whereas the generalization error can be large. 
The numerical results in Section~\ref{sec:numerical} further verify that this observation.

By \cite[Thm. 2]{smith_cocoa_nodate}
and strong duality of \eqref{eqn:problem},
the solution produced by Algorithm~\ref{alg:cocoa} is optimal for the optimization problem in  \eqref{eqn:problem}.
This is realized by making use of the concept ``bounded support modification'' in \cite[eqn. (18)]{smith_cocoa_nodate}.
We note that with $\lambda=0$ and $\sigma_w^2=0$,
an optimal solution gives exactly zero training error \eqref{eqn:training_error}.
On the the other hand, 
there can exist multiple solutions with zero training error,
but with vastly varying generalization error.
Hence,
a characterization of the generalization error of the distributed algorithm is needed, 
which is the focus of this paper. 

Motivated by the 
recent results on overparameterization in linear regression \cite{nakkiran2020optimal,bartlett2020benign, hastie2020surprises, Sahai_Harmless}
and the success of massively overparameterized models \cite[Table 1]{zhang_understanding_2017},
we pay special attention to the overparameterized setting of $p_k\geq n$, $\forall k$.
Although our results generally hold for all possible partitioning schemes,
we obtain some particularly interesting results for this overparameterized setting.
The following lemma shows that the governing matrix $\Bmat$  is a projection under $p_k\geq n$, $\forall k$, which will be instrumental for the upcoming results:
    \begin{lemma}\label{lemma:b_proj}
        Let $\Bmat$ be defined as in \eqref{eqn:xhattp:B}, 
        and the rows of $\Amat$ be drawn i.i.d. from $\Nc(0,\Sigmamat)$,
        with $\Sigmamat \in \mathbb{R}^{p \times p}$ being positive definite.
        If all $\Amat_k$ are broad, i.e.,
        $p_k \geq n$, $\forall k$, 
        then with probability one, we have $\Bmat^2=\Bmat$.
    \vspace*{2pt}
    \end{lemma}

\noindent Proof: See Section~\ref{sec:lemma:b_proj}. 

Note that Lemma~\ref{lemma:b_proj} together with \eqref{eqn:xhattp:x} shows that \cocoa{}, i.e., Algorithm~\ref{alg:cocoa}, converges in one iteration if the number of unknowns at each node is larger or equal to the number of training samples,
i.e., $\xhat^t = \xhat^1$ for $t\geq 1$ if $p_k\geq n, \forall k$.
Hence we have the following corollary for the generalization error:
\input{props/col_generr_broad_setting}

\noindent Proof: Using Lemma~\ref{lemma:b_proj}, we have that 
    $\Bmat^t = \Bmat$ 
    and 
    $\Rmat_t = \frac{1}{K}\Abar$,
    hence $\xhatt=\xhat^1$. 
    Combining these equations with 
    \eqref{eqn:noisy_egen_Bbound:termstogether} 
    gives the expression in \eqref{eqn:generr_broad_setting_noisy}.
    Similarly as with 
    \eqref{eqn:noisy_egen_Bbound} and \eqref{eqn:egen_Bbound},
    \eqref{eqn:generr_broad_setting_noisy} can be upper bounded by \eqref{eqn:generr_broad_setting_noisy_b}, 
    and 
    by \eqref{eqn:generr_broad_setting}
    if $\wvec=0$.
    \qed

Note that Lemma~\ref{lemma:b_proj}
provides an interesting observation for the training error in \eqref{eqn:training_error}.
Using that $\Bmat^t = \Bmat$ for $t\geq 1$,
we can simplify the training error in \eqref{eqn:training_error}:
\begin{align}\label{eqn:training_error_broad_case}
    \frac{1}{n}\|\Amat \Bmat \xvec\|^2 
    & =  \frac{1}{n}\|\Amat(\eye{p} - \frac{1}{K}\Abar \Amat)\xvec\|^2 \\
    & =  \frac{1}{n}\|(\Amat - \frac{1}{K}\Amat \Abar \Amat )\xvec\|^2 = 0,
\end{align}
where we have used that $p_k\geq n$ to apply
$\Amat\Abar = K\eye{n}$,
using
Property~\ref{prop:gauss_identity}
from Section~\ref{sec:preliminaries} of the Appendix.
Our results in Theorem~\ref{theorem:Bbound_prob} -- \ref{thm:subg_2} of the coming sections illustrate that $\|\Bmat\|$ can be unboundedly large if $p_k$ is too close to $n$.
Hence the generalization error in \eqref{eqn:generr_broad_setting} can be unbounded even though the training error is zero.

We conclude this section by motivating our study of the unregularized case.
With $\lambda>0$, convergence to a solution with arbitrarily small optimality gap is guaranteed with a sufficient number of iterations $T$ \cite[Thm.~3]{smith_cocoa_nodate}.
Hence in our problem setting, for $T$ large enough
we have that $\xhat^T\rightarrow (\Amat\T\Amat+\lambda\eye{p})\inv\Amat\T\yvec$, i.e., convergence to the centralized least-squares (LS) solution in  \eqref{eqn:centralized_sol}.   
While \cite[Thm.~3]{smith_cocoa_nodate} 
shows that smaller $\lambda$ will require larger $T$,
these results do not show how the partitioning affects the generalization error.
We address this gap by first studying the setting with $\lambda=0$ analytically,
and then showing implications of these results for the setting with $\lambda>0$ before convergence through numerical results.

\input{gen_error_iid_gauss}

\input{gen_error_cov_gauss}

\input{gen_error_sub_gauss}

%% file: props/col_generr_broad_setting.tex
\begin{corollary}
\label{col:generr_broad_setting}
    Consider the setting of Lemma~\ref{lemma:b_proj},
    and $t\geq 1$.
    Then $\xhatt = \xhat^1$, and
    \begin{align}\label{eqn:generr_broad_setting_noisy}
        \!\!
        \egen(\xhat^t) 
        = \egen(\xhat^1) 
        &=  \|\Sigmamat^{1/2}(
            \Bmat \xvec - \frac{1}{K}\Abar \wvec
            )\|^2 \\
        &\leq 2\|\Sigmamat\|\big(
            \|\Bmat\|^2\|\xvec\|^2
            \!+\! \frac{1}{K^2}\|\Abar\|^2\|\wvec\|^2\big),\label{eqn:generr_broad_setting_noisy_b}
    \end{align}
    which for $\wvec=0$ can be tightened as
    \begin{equation}\label{eqn:generr_broad_setting}
        \egen(\xhat^t)\leq \|\Sigmamat\| \|\Bmat\|^2 \|\xvec\|^2.
    \end{equation}
    
\end{corollary} 

%% file: gen_error_iid_gauss.tex
\section{Isotropic Gaussian Regressors}\label{sec:gaussian:iid}
In this section we present our analysis on the generalization error associated with Algorithm \ref{alg:cocoa} under isotropic Gaussian regressors,
i.e., the entries of $\Amat$ are i.i.d. with $\Nc(0,1)$, 
or equivalently, the rows of $\Amat$ are i.i.d. with ${\avec_i\sim\Nc(0,\eye{p})}$.
In Section~\ref{sec:gaussian:correlated} and Section~\ref{sec:subgaussian},  we extend our results to the correlated Gaussian and the sub-gaussian settings.
Focusing first on the isotropic Gaussian distributions allows us to give more precise results than the case for more general distributions, see the discussions after Remark~\ref{remark:corr_gaussian} and the discussions at the end of Section~\ref{sec:subgaussian} for details.
    
\input{props/lemma_tracy_widom}
This result quantifies the deviations of the extreme singular values of a standard Gaussian random matrix from their respective expectations.
Our main result in this section uses Lemma~\ref{lemma:tracy_widoms_wisdom} to provide high-probability bounds for the spectral norm of $\| \Bmat \|$:

\input{props/theorem_Bbound_prob}
\noindent Proof:
See Section \ref{proof:theorem_Bbound_prob}.
Note that the probabilistic result in Theorem~\ref{theorem:Bbound_prob} is with respect to the distribution of the training data in $\Amat$,
whereas the expectation taken in $\egen(\xhatt)$,
see \eqref{eqn:test:crossvanish},
is with respect to the unseen data, 
i.e., the test regressors.

Theorem~\ref{theorem:Bbound_prob} provides key insights about  $\Bmat$,
which governs the iterations of \cocoa{},
see \eqref{eqn:xhattp:x}.
The matrix $\Bmat$ represents the contribution of the local solutions $\xhat_k$ from each node,
as well as the interactions of these solutions through the shared variable $\bar{\vvec}$.
In one node at a given iteration, 
the setting can be interpreted as a regression problem with a partial model with missing features, 
i.e., that some of the features in the full model are ignored during regression. 
The main technical challenge in Theorem~\ref{theorem:Bbound_prob} is then to capture the combined contribution of many nodes to the error,
each of which sees its own partial model. 
In \eqref{eqn:Bbound_upper_prob},
the bound $\beta_G$ on the spectral norm of $\Bmat$ captures this joint behaviour,
with dependency on the dimensions of the local regressor matrices $\Amat_k$.
    
We now connect the presented bound on $\|\Bmat\|$ to the generalization error $\egen(\xhatt)$,
illustrating how the overall performance of the solution can depend on the partitioning scheme.
Together with \eqref{eqn:egen_Bbound}, Theorem~\ref{theorem:Bbound_prob} provides an upper bound on the generalization error:
\begin{align}\label{eqn:ebound:Gaussian:iid}
    \egen(\xhatt) \leq \beta_G^{2t} \| \xvec \|^2,
\end{align}
with success probability, i.e., probability of the upper bound holding, at least $\rho_G$.
Ideally, the bound in \eqref{eqn:Bbound_upper_prob} would be small while the success probability $\rho_G$ is large, meaning that the generalization error is small with a high probability.
To have a high success probability $\rho_G$, 
the variables $q_i$ needs to be large for $\ikrange$.
On the other hand, 
as any $q_i$ approaches its upper bound, 
i.e., $q_i\rightarrow \sqrt{\rmaxi} - \sqrt{\rmini}$, 
the corresponding denominator in $\beta_G$ goes to zero,
and the upper bound becomes larger.
Thus, we need $\sqrt{\rmaxi} - \sqrt{\rmini}$ to be sufficiently large for $\ikrange$,
so that all $q_i$ can be chosen to guarantee a sufficiently large $\rho_G$, 
without compromising the level of the upper bound.
Note that for $\rmaxi \approx \rmini$, Lemma~\ref{lemma:tracy_widoms_wisdom} is also typically uninformative. 
Hence, 
for a fixed success probability $\rho_G$, 
the bound on $\|\Bmat\|$ grows as $\rmaxi$ and $\rmini$ get closer.
Although results that provide a more accurate picture of the behaviour of the minimum singular value of $\Amat_i$ for 
$\rmaxi \approx \rmini$ \cite[Thm.~3.3]{rudelson_non-asymptotic_2010} exist,
a similar line of argument in terms of the effect of  $\rmaxi \approx \rmini$ also holds there.

\begin{remark}\label{remark:isoG}
    Theorem~\ref{theorem:Bbound_prob} shows that the bound $\beta_G$ guarantees smaller values on the generalization error with higher probability when the dimensions of the submatrices $n$ and $p_k$ are apart from each other,
    compared to when they are close. 
    This sufficiency result suggests that it may be the case that the generalization error gets large values with $n \approx p_k$. 
    \vspace*{2pt}
\end{remark}

Our results on the average generalization error in Section~\ref{sec:gauss:iid:expected}
and also the numerical results in Section~\ref{sec:numerical}
show that this is indeed the case.
Theorem~\ref{theorem:Bbound_prob} is consistent with the results for the isotropic Gaussian regressors for the centralized setting in
\cite{belkin_two_2019, breiman_how_nodate, belkin_reconciling_2019},
where the relationship between the number of unknowns and the number of observations determines whether a low generalization error is attainable. 
In particular,
average behaviour of the generalized inverse of Wishart  matrices \cite{belkin_two_2019, breiman_how_nodate},  as well as other high probability results  \cite{belkin_two_2019}
play a central role leading to ``double descent'' curves \cite{belkin_two_2019}. 
Similarly in Theorem~\ref{theorem:Bbound_prob},
the spectral properties of the Wishart matrices $\Amat_k \Amat_k\T$, 
particularly their singular values' closeness to zero, 
are of central importance. 
An important difference between the existing literature and our work in this section (as well as our work in the cases of correlated Gaussian and sub-gaussian distributions in the subsequent sections) is the fact that our results focus on the distributed setting and we explain how the trade-offs between the number of observations and the unknowns studied in the centralized case
have important implications for distributed learning.

For the setting where all local models at the nodes are overparameterized, 
we present the following tighter bound that holds for all $t$ on the generalization error:
\begin{corollary}\label{cor:pklarge:gaussiid:highprobability}
    Let $p_k \geq n, \forall k$. The generalization error $\egen(\xhatt)$ is bounded by $\egen(\xhatt) \leq \beta_G^{2} \| \xvec \|^2 $ with probability $\rho_G$. 
\end{corollary}

\noindent 
    Proof: The result follows directly from  Theorem~\ref{theorem:Bbound_prob} and 
    Corollary~\ref{col:generr_broad_setting}.
\qed

We now provide the following alternative bound for $\|\Bmat\|$ for the case with $n\geq p_k$, $\forall k$:
\input{props/lemma_alternative_bbound}
\noindent Proof: See Section~\ref{proof:lemma_alt_bound}.
This result provides an alternative to \eqref{eqn:ebound:Gaussian:iid}:
$
    \egen(\xhatt) \leq \Bar{\beta}_G^{2t} \|\xvec\|^2,
$
which holds with probability at least $\Bar{\rho}_G$.
In Section~\ref{sec:expectationvsprobability}
we compare the high-probability results of this section, in particular Corollary~\ref{cor:pklarge:gaussiid:highprobability} and Lemma~\ref{lemma:alternative_bound},
to the average generalization error.

\input{gen_error_conference_results}

%% file: props/lemma_tracy_widom.tex
\begin{lemma}\emph{(Tracy-Widom fluctuations \cite{rudelson_non-asymptotic_2010}).}
\label{lemma:tracy_widoms_wisdom}
For a matrix $\Mmat\in\Rbb^{n\times p}$ with i.i.d. $\Nc(0,1)$ distributed entries,
the following bound holds with probability at least $\rho = 1-2e^{-q^2/2},~{q\geq 0}$
    \begin{align}\label{eqn:tracy_widom}
        \begin{split}
            &\sqrt{r_{\max}} - \sqrt{r_{\min}} - q 
            \leq \sigma_{\min}(\Mmat) \\
            &\leq \sigma_{\max}(\Mmat)
            \leq \sqrt{r_{\max}} + \sqrt{r_{\min}} + q,
        \end{split}
    \end{align}
    where $r_{\min} = \min\{n,p\}$ and 
    $r_{\max}=\max\{n,p\}$.
    
\end{lemma}

%% file: props/theorem_Bbound_prob.tex
\begin{theorem}\label{theorem:Bbound_prob}
Let $\Amat\in\Rbb^{n\times p}$ be a Gaussian random matrix with i.i.d. $\Nc(0,1)$ distributed entries,
and let $\Bmat$ be defined as in \eqref{eqn:xhattp:B}. Let us define $r_{\min, k}=\min\{p_k,n\}$ and $r_{\max, k}=\max\{p_k,n\}$, and let 
\begin{align}\label{eqn:def:betaG}
    \beta_G \,
    {=} 1\, {+} \, \frac{1}{K}\!\sqrt{K {+}  \sum_{k=1}^K\sum_{\substack{i=1\\i\neq k}}^K \Bar{\gamma}_{k,i}
        },
\end{align}
where     
\begin{equation} \label{eqn:def:bargammaki}
      \Bar{\gamma}_{k,i} = \frac{(\sqrt{\rmaxk} + \sqrt{\rmink} + q_k)^2}
            {(\sqrt{\rmaxi}-\sqrt{\rmini} - q_i)^2},
\end{equation}
and ${0\leq q_i < \sqrt{r_{\max,i}} - \sqrt{r_{\min,i}}}$, $i=1,\,...,\,K $.
Then, the following bound
\begin{align}\label{eqn:Bbound_upper_prob}
    \|\Bmat\|
    \leq \beta_G,
\end{align}
holds with probability at least 
${\rho_G = \prod_{k=1}^K{(1-2e^{-q_k^2/2})_+}}$.

\end{theorem}

%% file: props/lemma_alternative_bbound.tex
\begin{lemma}\label{lemma:alternative_bound}
    Consider the setting of Theorem~\ref{theorem:Bbound_prob} under ${n\geq p_k},\forall k$.
    Let 
    \begin{equation}
        \Bar{\beta}_G = \sqrt{\frac{(K-1)^2}{K} 
            + \frac{1}{K^2}
            \sum_{k=1}^K
            \sum_{\substack{i=1\\i\neq k}}^K 
            \Bar{\gamma}_{k,i}
            },
    \end{equation}
    with $\Bar{\gamma}_{k,i}$ as defined in Theorem~\ref{theorem:Bbound_prob}.
    Then, the following bound
    \begin{equation}\label{eqn:lemma:alternative_bound}
       \|\Bmat\| \leq \Bar{\beta}_G,
    \end{equation}
    holds with probability at least
    $\Bar{\rho}_G = \prod_{k=1}^K (1-2e^{-q_k^2/2})_+$.
    
\end{lemma}

%% file: gen_error_conference_results.tex
\kern-1em
\subsection{The Average Generalization Error}\label{sec:gauss:iid:expected}
\kern-0.25em
We now compare our results with the average generalization error
    \setlength{\abovedisplayskip}{1pt}
    \setlength{\belowdisplayskip}{2pt}    
    \begin{align}\label{eqn:generalization_error_def}
        \Ebb_{\Amat,\wvec} [ \egen(\xhat)] &=
        \Ebb_{\Amat,\wvec} [ \| \xvec-\xhat\|^2],
    \end{align}
    \setlength{\abovedisplayskip}{4pt}%
    \setlength{\belowdisplayskip}{4pt}%
where the expectation is over the regressor matrix $\Amat$ in the training data and the training noise $\wvec$.
We consider 
the following extension of \cite[Thm.~1]{HellkvistOzcelikkaleAhlen_distributed2020_spawc} to the case with training noise:
    
\input{props/lemma_conference_result}

\noindent Proof: See Section~\ref{sec:proof:conference_result}.
Similar to Theorem~\ref{theorem:Bbound_prob}, 
the average error diverges for  $p_i\in\{n\!-\!1,n,n\!+\!1\}$ due to pseudo-inverses, 
see \cite{HellkvistOzcelikkaleAhlen_distributed2020_spawc} and the related discussions in the longer version \cite[Section~VII-C]{HellkvistOzcelikkaleAhlen_distributed2020_technicalReport}. 
\begin{remark}
    Both Theorem~\ref{theorem:Bbound_prob} and  Lemma~\ref{lemma:conference_result} suggest that we may have a large generalization error when the number of unknowns at a node is close to the number of observations, i.e.,
    at least one of the local system of equations is approximately square,
    regardless of being under- or over-parameterized. 
     \vspace*{2pt}
\end{remark}

The following corollary illustrates how Lemma~\ref{lemma:conference_result} can be used to provide an error expression for all iterations in the overparametrized case: 
\begin{corollary}\label{cor:pklarge:gaussiid:expectation}
    Let $p_k \geq n, \forall k$.
    The average generalization error for any iteration $t \geq 1$ is given by 
    \begin{align}\label{eqn:xtildetp:pklarge}
        \Ebb_{\Amat,\wvec} [ \egen(\xhatt)] 
        = \sum_{k=1}^K \norm{\xvec_k }^2 \alpha_k 
            + \frac{\tr(\Sigmamat_w)}{K^2}\gamma_k,
    \end{align}
    where $\alpha_k$, $\gamma_k$ are given by Lemma~\ref{lemma:conference_result}. 
    
\end{corollary}
\noindent 
Proof: 
By combining Lemma~\ref{lemma:conference_result} and Corollary~\ref{col:generr_broad_setting}, i.e.,
$\egen(\xhatt) = \egen(\xhat^1)$, we obtain the desired result. 
\qed

The numerical results of \cite{HellkvistOzcelikkaleAhlen_distributed2020_spawc} as well as the results in Section~\ref{sec:numerical} (see Figure~\ref{fig:final_mse}) suggest that \eqref{eqn:xtildetp:pklarge} provides not only the error for the overparametrized case but also reveals the general approximate behaviour of the algorithm even if the $p_k \geq n$ condition is not satisfied for $\krange$. 

\input{props/example_iso_G}

%% file: props/lemma_conference_result.tex
\begin{lemma}%
\label{lemma:conference_result}
    Let $\Amat\in\Rbb^{n\times p}$ be a random matrix with i.i.d. $\Nc(0,1)$ distributed entries.
    Let $\wvec\in\Rbb^{n\times 1}$ be a zero-mean random vector with covariance matrix $\Sigmamat_w$,
    statistically independent with the regressors. 
    The average generalization error in iteration $t=1$ of Algorithm~\ref{alg:cocoa}, can be expressed as
    \begin{equation}\label{eqn:xtildetp}
         \Ebb_{\Amat,\wvec} [ \egen(\xhat^1)]  
         = \sum_{k=1}^K \norm{\xvec_k }^2 \alpha_k
        + \frac{\tr(\Sigmamat_w)}{K^2}\gamma_k,
    \end{equation}
    where $\alpha_k,~\gamma_k, \,k=1,\,\dots,K$, are given by
            \begin{equation}\textstyle\label{eqn:alpha_k}
                \hspace{-47pt}\alpha_k 
                = \frac{1}{K^2}(
                    K^2 
                    + (1-2K)\tfrac{r_{\min,k}}{p_k} 
                    + \sum_{\substack{i=1\\i\neq k}}^K\gamma_i
                ),
            \end{equation}
            \begin{subnumcases}{\label{eqn:gamma_main}\hspace{-10pt} \gamma_k=}
                \tfrac{r_{\min, k}}{r_{\max, k} - r_{\min, k} - 1} & \hspace{-5pt}for $p_k \notin \{n-1, n, n+1\},$ \label{eqn:gamma_main_a}\\
                +\infty & \hspace{-15pt}otherwise, \label{eqn:gamma_main_b}
            \end{subnumcases}
    and $r_{\min, i}=\min\{p_i,n\}$ and $r_{\max, i}=\max\{p_i,n\}$.
    
\end{lemma}

%% file: props/example_iso_G.tex
\kern-1em
\subsection{Comparison with the Average Generalization Error}
\kern-0.25em
    \label{sec:expectationvsprobability}
    We now consider an example where we first study the expectation results from Lemma \ref{lemma:conference_result} and Corollary \ref{cor:pklarge:gaussiid:expectation},
    and then compare them to the probabilistic results in Lemma~\ref{lemma:alternative_bound}  and Corollary~\ref{cor:pklarge:gaussiid:highprobability}.
    We here consider the setting with $\wvec=0$.
    
    Let $\| \xvec_k \|^2 =\frac{1}{K}$. 
    By Lemma~\ref{lemma:conference_result}, 
    we have $\Ebb_{\Amat}[ \egen(\xhat^1)] =  \frac{1}{K} \sum_{k=1}^K  \alpha_k $.
    Using $\frac{\rmink}{p_k} \leq 1$,
    we obtain
    \begin{align}
        \Ebb_{\Amat}[ \egen(\xhat^1)] 
        &\leq \frac{1}{K} 
            \left( 
                \frac{(K-1)^2}{K} 
                + \frac{1}{K^2}\sum_{k=1}^K\sum_{\substack{i=1\\i\neq k}}^K \gamma_i
            \right), \label{eqn:xknormsequal:semi} \\
        &\leq
        1 + \frac{1}{K^2}
        + \frac{1}{K^3}\sum_{k=1}^K\sum_{\substack{i=1\\i\neq k}}^K \gamma_i.\label{eqn:xknormsequal:expectation}
    \end{align}
    Under $p_k\geq n$, using Corollary \ref{cor:pklarge:gaussiid:expectation} we observe that
    \begin{align}\label{eqn:E_xhatt=xhat1}
        \Ebb_{\Amat}[ \egen(\xhat^t)] = 
        \Ebb_{\Amat}[ \egen(\xhat^1)].
    \end{align}
    
    We now consider the probabilistic results in two separate cases:
    
    i) Let $n\geq p_k,\forall k$. By Lemma~\ref{lemma:alternative_bound}, the following holds with probability at least $\Bar{\rho}_G$ for $t=1$: 
    \setlength{\abovedisplayskip}{1pt}
    \setlength{\belowdisplayskip}{1pt}
    \begin{equation}\label{eqn:egen_1_tall}
        \egen(\xhat^1) \leq \Bar{\beta}_G^2 = \frac{(K-1)^2}{K} 
                + \frac{1}{K^2}\sum_{k=1}^K\sum_{\substack{i=1\\i\neq k}}^K \Bar{\gamma}_{k,i}.
    \end{equation}
    \setlength{\abovedisplayskip}{4pt}%
    \setlength{\belowdisplayskip}{4pt}%
    Comparing \eqref{eqn:egen_1_tall} with \eqref{eqn:xknormsequal:semi}, 
    we observe that the expressions have a shared algebraic form where the expectation result in \eqref{eqn:xknormsequal:semi} has a scaling of $\frac{1}{K}$ compared to the probabilistic result in \eqref{eqn:egen_1_tall},
    under $\Bar{\gamma}_{k,i} = \gamma_i$.
    Both results reveal how the dimensions of the local data matrices $\Amat_k$ affect the error:  the expectation results in \eqref{eqn:xknormsequal:semi}  through $\gamma_i$ and the probability results in \eqref{eqn:egen_1_tall} through $\bar{\gamma}_{k,i}$.
   
    ii) Let $p_k \geq n,\forall k$. 
    Using Corollary~\ref{cor:pklarge:gaussiid:highprobability} and \eqref{eqn:def:betaG},
    we observe that the following holds with probability at least $\rho_G$
    \begin{align}
        \egen(\xhat^t) 
        \leq \beta_G^2 
        \leq 2 + \frac{2}{K} + \frac{2}{K^2} \sum_{k=1}^K\sum_{\substack{i=1\\i\neq k}}^K \bar{\gamma}_{k,i}, 
        \label{eqn:xknormsequal:highprob}
    \end{align} 
    where we used $(a+b)^2 \leq 2 a^2 +2 b^2 $ on \eqref{eqn:def:betaG}. 
    Using \eqref{eqn:E_xhatt=xhat1},
    we compare \eqref{eqn:xknormsequal:highprob} to \eqref{eqn:xknormsequal:expectation}: 
    The two bounds again quantify the dependence of the error on the partitioning  through $\gamma_i$ and $\bar{\gamma}_{k,i}$ 
    and they have the same shared form under $\gamma_i=\bar{\gamma}_{k,i}$.
    
    This example emphasizes the common algebraic structure in the expectation and the high-probability results.
    Although the bounds in \eqref{eqn:egen_1_tall}/\eqref{eqn:xknormsequal:highprob} and \eqref{eqn:xknormsequal:semi}/\eqref{eqn:xknormsequal:expectation}
    contain different constant additive terms, they all heavily depend on the terms $\bar{\gamma}_{k,i}$ and $\gamma_{i}$ which are the main factors characterizing the behaviour of the generalization error with respect to the partitioning.

%% file: gen_error_cov_gauss.tex
\section{Correlated Gaussian Regressors}\label{sec:gaussian:correlated}
This section generalizes the results of the preceding section to correlated Gaussian regressors.
The regressors $\avec_i$ (i.e., the rows of $\Amat$) are now i.i.d. zero-mean random vectors drawn from the Gaussian distribution $\Nc(0,\Sigmamat)$,
with $\Sigmamat = \Ebb_{\avec_i}[\avec_i\avec_i^T]$,  
i.e., each row of $\Amat$ is independently drawn from  $\Nc(0,\Sigmamat)$. 
Our main result in this section is given by Theorem~\ref{theorem_correlated_Bbound_prob}:

\input{props/theorem_correlated_Bbound_prob}
\noindent Proof: See Section~\ref{proof:theorem_correlated_Bbound_prob}.
Similar to  Theorem~\ref{theorem:Bbound_prob},
Theorem~\ref{theorem_correlated_Bbound_prob} illustrates the connection between the dimensions of the local matrices $\Amat_k$ and the norm $\|\Bmat\|$.

We also present a result for the special case $n\geq p_k$,
analogous to Lemma~\ref{lemma:alternative_bound} but for the correlated Gaussian setting:
\kern-0.5em
\begin{lemma}\label{lemma:alternative_bound_corr_gauss}
    Consider the setting of Theorem~\ref{theorem_correlated_Bbound_prob},
    under $n\geq p_k,\forall k$.
    With probability at least 
    $1-2 \sum_{k=1}^K e^{-q_k}$,
    \eqref{eqn:lemma:alternative_bound} holds with the following redefinition
    $\Bar{\gamma}_{k,i}=(n\sigmamax(\Sigmamat_k) + \ell_k(q_k))/(n\sigmamin(\Sigmamat_k) - \ell_i(q_i))_+$.
    
\end{lemma}
\noindent Proof:
The proof follows the same line of argument as Lemma~\ref{lemma:alternative_bound},
where $\Bar{\gamma}_{k,i}$ is defined using \eqref{eqn:thm:corr_Bbound_beta_1}. \qed

We now combine Theorem \ref{theorem_correlated_Bbound_prob} 
with \eqref{eqn:egen_Bbound} and obtain the following upper bound on the generalization error  
\begin{equation}
    \egen(\xhatt) \leq \beta_{G_c}^{2t}\|\Sigmamat\|\|\xvec\|^2.
\end{equation}

\begin{remark}\label{remark:corr_gaussian}
    Theorem \ref{theorem_correlated_Bbound_prob} is
    consistent with Theorem \ref{theorem:Bbound_prob},
    also illustrating how the generalization error of the solution produced by \cocoa{} (Algorithm~\ref{alg:cocoa}) is affected by the partitioning scheme: if $n$ and $p_k$ are sufficiently far apart  and
    $\Sigmamat_k$'s are well-conditioned, similar to the case of $\Sigmamat =\eye{p}$,
    then a low generalization error  is guaranteed with high probability.
    \vspace{2pt}
\end{remark}

We note that extending the results from
the isotropic to the correlated Gaussian 
setting introduced more complexity in the expressions.
In Theorem~\ref{theorem_correlated_Bbound_prob},
the bound on $\|\Bmat\|$ is expressed up to an absolute constant $C$,
whereas in the isotropic setting of Theorem~\ref{theorem:Bbound_prob},  a more refined bound was presented based on the stronger results for typical behaviour of such Gaussian matrices, as in Lemma~\ref{lemma:tracy_widoms_wisdom}.

Theorem~\ref{theorem_correlated_Bbound_prob} emphasizes the relation between the partitioning and the generalization error, and points out a nontrivial dependency on the regressors' covariance matrix 
through the dependence on $\sigmamin(\Sigmamat_k)$ and $\sigmamax(\Sigmamat_k)$. 
These results are consistent with the results in the centralized setting \cite{hastie2020surprises,bartlett2020benign}, \cite[Section 5.1]{nakkiran2020optimal},
which also illustrate that 
the performance is connected  to both the dimensions of the problem as well as the spectral properties of $\Sigmamat$.
In \cite{hastie2020surprises} and \cite[Section 5.1]{nakkiran2020optimal},
it is emphasized that the generalization performance depends not only on the dimensions of the problem,
but also on the relative geometry of the regressors' covariance matrix and the unknowns.
In \cite{bartlett2020benign}, the decay of the covariance matrix's singular values is emphasized as a key indicator of whether a small generalization error can be achieved with sub-gaussian regressors, for which Gaussian regressors is a special case.
An important point here is the distinction between the error in $\xvec$ and the generalization error.
For instance, in \eqref{eqn:thm:corr_Bbound_beta_2}, a large discrepancy between $\sigmamax(\Sigmamat_k)$ and $\sigmamin(\Sigmamat_k)$ will lead to a large bound on $\|\Bmat\|$ for  $p_k > n$ whereas by \eqref{eqn:egen_iter_t} the generalization error can be potentially small, for instance, when there is only one large eigenvalue, hence typical regressor realizations are approximately the same; and hence the generalization error (error in $\avec\T \xvec$) is small.

%% file: props/theorem_correlated_Bbound_prob.tex
\kern-0.5em
\begin{theorem}\label{theorem_correlated_Bbound_prob}
    Let $\Bmat$ be defined as in \eqref{eqn:xhattp:B},
    and the rows of $\Amat$ be i.i.d. with $\Nc(0,\Sigmamat)$,
    $\Sigmamat\succ 0$,
    and let $\Kc = \{k:n<p_k\}$.
    Let
    \begin{equation}\label{eqn:thm_correlated_Bbound_prob_upper}
        \beta_{G_c} = 
        1 + \frac{1}{K}
        \sqrt{K + \sum_{k=1}^K\sum_{\substack{i=1\\i\neq k}}^K
            \frac{
                n\sigmamax(\Sigmamat_k) + \ell_k(q_k)
            }{
                \eta_i
            }
        },
    \end{equation}
    where $\Sigmamat_k$ is the $k$\th{} principal submatrix of $\Sigmamat$ and
    \begin{subnumcases}{\label{eqn:corr_gauss_cases_main} \hspace{-20pt} \!\eta_k=}
       \! (n\sigmamin(\Sigmamat_k) - \ell_k(q_k))_+,~n \geq p_k 
       \label{eqn:thm:corr_Bbound_beta_1} \\
       \! \sigmamin(\Sigmamat_k)
       (\sqrt{p_k}\! -\! \sqrt{n} - \Bar{q}_k)_+^2,~ n < p_k
       \label{eqn:thm:corr_Bbound_beta_2}
    \end{subnumcases}
    and
    \begin{equation}
        \ell_k(q_k) = \tfrac{8}{3}\, n\, \sigmamax(\Sigmamat_k) \, C 
        \left(
            \sqrt{\tfrac{p_k+q_k}{n}} + \tfrac{p_k+q_k}{n}
        \right),
    \end{equation}
    where $C$ is an absolute constant, 
    and $q_k,\,\Bar{q}_k\geq 0$,
    $\,\forall k$.
    Then, the following bound
    \begin{equation}\label{eqn:bbound_beta_Gc_corr}
        \|\Bmat\| \leq \beta_{G_c},
    \end{equation}
    holds with probability at least 
    $\rho_{G_c}=1-\sum_{k=1}^K 2e^{-q_k} - \sum_{k\in \Kc} 2e^{-\Bar{q}_k^2/2}$.
    
\end{theorem}

%% file: gen_error_sub_gauss.tex
\section{Sub-gaussian Regressors}\label{sec:subgaussian}
In this section, we consider regressors drawn from sub-gaussian distributions. 
The family of sub-gaussian distributions include the Gaussian, uniform and the Bernoulli random variables as well as any other bounded random variable \cite{vershynin2018high}, 
hence it allows us to investigate a large range of data distributions.

\input{sub_gaussians}

\kern-0.05em
\subsection{Generalization Error under Sub-gaussian Regressors}\label{sec:subgaussian:thms}
\kern-0.05em

We now present our main results for the sub-gaussian case.
In the following theorem, we assume that the submatrices $\Amat_k\in\Rbb^{n\times p_k}$ are generated from a matrix $\Zmat_k\in\Rbb^{n\times p_k}$ where each entry of $\Zmat_k$ is drawn i.i.d. from $\Sc(1)$, $\forall k$.
This way of generating $\Amat_k$'s renders the matrices $\Amat_k$ statistically independent, 
and the covariance matrix of the rows of $\Amat$  block-diagonal.%

\input{props/theorem_sub_gauss_1}
\noindent Proof: See Section~\ref{proof:thm:sugb_1}.
Note that we use the subscript $i,k$ on $\avec_{i,k}$ in \eqref{eqn:blockdiagonalsubgauss:psi} to emphasize that $\avec_{i,k}$ is i.i.d. with the rows of $\Amat_k$.
In our results with Gaussian regressors, we utilize the fact that for the partitions $\Amat_k$ with i.i.d. Gaussian rows, 
there is always a decomposition with $\Zmat_k$ which has entries from $\Nc(0,1)$ (but $\Zmat_k$'s are not necessarily i.i.d.).
With sub-gaussian rows,
this type of inverse relationship (i.e. from $\Amat$ with sub-gaussian rows with a certain sub-gaussian norm to  $\Zmat_k$ with i.i.d. sub-gaussian elements with a given norm) is not straightforward. 
Hence, 
we here focus on covariance structures enabling such a relationship,
constructing $\Amat_k=\Zmat_k\Lambdamat_k^{1/2}\Umat_k\T$ in Theorem~\ref{thm:subg_1},
and assuming $n\geq p_k$ in the following theorem,

\input{props/theorem_sub_gauss_2}

\noindent Proof: See Section~\ref{proof:thm:subg_2}.

Similar to the previous results with Gaussian regressors, we obtain the bounds on the generalization error as
$
\egen(\xhatt) \leq \beta_S^{2t} \|\Sigmamat\| \|\xvec\|^2
$ 
and
$
\egen(\xhatt) \leq \beta_{S_c}^{2t} \|\Sigmamat\| \|\xvec\|^2,
$
by Theorem~\ref{thm:subg_1} and Theorem~\ref{thm:subg_2}, respectively.

Theorem~\ref{thm:subg_1} and \ref{thm:subg_2} provide analogous insights as Theorem~\ref{theorem:Bbound_prob} and \ref{theorem_correlated_Bbound_prob}
in the sense that the bounds $\beta_S$ and $\beta_{S_c}$ depend on the dimensions of the local regressor matrices $\Amat_k$ and on the corresponding covariance matrices $\Sigmamat_k$.

\begin{remark}\label{remark:subg}
    Theorem~\ref{thm:subg_1} and \ref{thm:subg_2} are consistent with Theorem~\ref{theorem:Bbound_prob} and \ref{theorem_correlated_Bbound_prob}:
    all of these results provide bounds on the generalization error that can be guaranteed to have smaller values if $n$ and $p_k$ are further apart compared to the case when they are closer. 
\end{remark}

Theorem~\ref{thm:subg_1} is consistent with the centralized setting of \cite{bartlett2020benign} with sub-gaussian regressors.
In~\cite{bartlett2020benign},
the dimensions $p$ and $n$ as well as the spectral properties of the regressors' covariance matrix $\Sigmamat$ are pointed out as important factors determining the generalization error.
We correspondingly highlight the local dimensions $p_k$ and $n$; and the local covariance matrices $\Sigmamat_k$.
In \cite{Sahai_Harmless},
bounds on the generalization error with sub-gaussian regressors,
focusing on the effects of training noise,
are derived in the centralized setting.
While our results focus on the noise-free distributed setting,
the implications of the noisy interpolation results of \cite{Sahai_Harmless} are considered as an important line of future work.    

The family of sub-gaussian distributions includes a large variety of distributions, including Gaussian regressors of Section~\ref{sec:gaussian:iid}--\ref{sec:gaussian:correlated} and  Bernoulli regressors, popular in compressive sensing \cite{foucartRauhut_2013}. 
Furthermore, 
all bounded distributions are sub-gaussian distributions.
For instance, 
the regressors formed by random Fourier features \cite{RahimiRecht_2008},
used in various classification tasks and also studied in Section~\ref{sec:mnist_simulations},
are sub-gaussian since the magnitude of the elements of these regressors are bounded by~$1$.
We note that our results for the Gaussian settings are more refined than those in Theorem~\ref{thm:subg_1} and \ref{thm:subg_2},
due to the existence of more precise results for Gaussian distribution than for the broad family of sub-gaussian distributions.
On the other hand,
Theorem~\ref{thm:subg_1} and \ref{thm:subg_2} cover the very general setting of sub-gaussian distributions, 
and could be further specialized for other special cases of sub-gaussian distributions, such as for applications with Bernoulli regressors.

%% file: sub_gaussians.tex
\kern-1em
\subsection{Preliminaries on Sub-gaussian Random Variables}\label{sec:preliminaries:subgaussian}
\kern-0.25em
This section provides preliminaries on sub-gaussian random variables \cite{vershynin2018high}.

\begin{definition}{\emph{(Sub-gaussian random variables)}}\label{def:subg_rvar}
    A random variable $z\in\Rbb$ is called sub-gaussian if there exists a constant $L>0$ so that the following is satisfied
    \begin{equation}\label{eqn:def:subgaussian}
        \Ebb\left[e^{z^2/L^2}\right] \leq 2.
    \end{equation}
    The smallest $L$  defines the sub-gaussian norm
    $\| z \|_{\psi_2}$ as follows
    \begin{equation}\label{eqn:def:subgaussiannorm:singlevariable}
        \| z \|_{\psi_2} = \inf\{L>0:~ \Ebb\left[e^{z^2/L^2}\right] \leq 2\}.
    \end{equation}
    
\end{definition}
This definition can be extended to higher dimensions:
\begin{definition}{\emph{(Sub-gaussian random vectors)}}\label{def:subg_rvec}
    A random vector $\zvec\in\Rbb^{p\times 1}$
    is called sub-gaussian if for all $\hvec\in\Rbb^{p\times 1}$,
    $\zvec\T\hvec$ is a sub-gaussian random variable.
    \vspace{2pt}
\end{definition}

With a slight abuse of notation, we use $\avec\sim\Sc(\Sigmamat)$ to denote that the random vector $\avec$ comes from some zero-mean sub-gaussian distribution $\Sc$, and has the covariance matrix $\Sigmamat$. 
We introduce the following notation for the sub-gaussian norm
\begin{align}\label{eqn:def:subgaussvectornorm}
\psi_{\avec}(\Sigmamat,\hvec) = \|\hvec\T \avec\|_{\psi_2},
\end{align}
where $\hvec\in\Rbb^{p\times 1}$.

%% file: props/theorem_sub_gauss_1.tex
\kern-0.5em
\begin{theorem}\label{thm:subg_1}
    Let the matrix $\Bmat$ be defined as in \eqref{eqn:xhattp:B},
    with each $\Amat_k$ generated as $\Amat_k=\Zmat_k\Lambdamat_k^{1/2}\Umat_k\T\in\Rbb^{n\times p_k}$,
    where the entries of $\Zmat_k\in\Rbb^{n\times p_k}$ are i.i.d. with $\Sc(1)$, $\forall k$,
    $\Lambdamat_k\in \Rbb^{p_k\times p_k}$ is diagonal and positive definite, and $\Umat_k\in\Rbb^{p_k\times p_k}$ is unitary.
    Let $\Sigmamat_k $ denote the associated covariance matrix for the rows of $\Amat_k$. 
    Let $\beta_{S}$ be defined as
    \begin{equation}
        \beta_{S} = 1 + \frac{1}{K}\sqrt{
            K + \sum_{k=1}^K\sum_{\substack{i=1\\i\neq k}}^K
            \frac{n\sigmamax(\Sigmamat_k) + \ell_k(q_k)}{\eta_i}
        },
    \end{equation}
    where
    \begin{subnumcases}{\label{eqn:subg_gauss_cases_main} \hspace{-20pt}\!\eta_k=}
       \! (n\sigmamin(\Sigmamat_{k}) - \ell_k(q_k))_+,~n \geq p_k, \\
       \! \sigmamin(\Sigmamat_{k})
       (\sqrt{p_k}\! -CL_k^2(\! \sqrt{n} + \Bar{q}_k))_+^2,~n<p_k,\label{eqn:subg_gauss_cases_b}
    \end{subnumcases}
    \begin{equation}\label{eqn:blockdiagonalsubgauss:lk}
        \ell_k(q_k) = CL_k^2 
        \left(\sqrt{\frac{p_k+q_k}{n}} + \frac{p_k+q_k}{n}\right),
    \end{equation}
    and $C$ is an absolute constant, $q_k,\,\Bar{q}_k \geq 0$,
    $\,\forall k$,
    and $L_k\geq 1$ are constants such that, for all $\hvec\in\Rbb^{p_k\times 1}$
    \begin{equation}\label{eqn:blockdiagonalsubgauss:psi}
        \psi_{\avec_{i,k}}(\Sigmamat_k, \hvec) \leq L_k \sqrt{\hvec\T \Sigmamat_k \hvec},
    \end{equation}
    where $\avec_{i,k} \in \Rbb^{p_k\times 1}$ comes from the same distribution as the rows of $\Amat_k$.  
    Then, the following bound holds 
    \begin{equation}\label{eqn:B:thm:subg_1}
        \|\Bmat\| \leq \beta_S,
    \end{equation}
    with probability at least 
    $\rho_S 
    {=} 1 {-} \sum_{k=1}^K 2e^{-q_k} 
        {-} \sum_{k\in\Kc} 2e^{-\Bar{q}_k^2}$.
\end{theorem}

%% file: props/theorem_sub_gauss_2.tex
\kern-0.5em
\begin{theorem}\label{thm:subg_2}
    Let the matrix $\Bmat$ be defined as in \eqref{eqn:xhattp:B} and
    the rows of $\Amat$ be i.i.d. with $\Sc(\Sigmamat)$,
    with $n\geq p_k,\,\forall k$,
    and $\Sigmamat\succ 0$.
    Let $\Sigmamat_k$ denote the $k$\th{} principal submatrix 
    \cite[Sec. 0.7.1]{horn1990matrix} of $\Sigmamat$.
    Let $\beta_{S_c}$ be defined as
    \begin{equation}\label{eqn:Bbound_subg_2}
        \beta_{S_c} = 1 + \tfrac{1}{K}\sqrt{
            K + \sum_{k=1}^K\sum_{\substack{i=1\\i\neq k}}^K
            \frac{n\sigmamax(\Sigmamat_k) + \ell_k(q_k)}{(n\sigmamin(\Sigmamat_i) - \ell_i(q_i))_+}
        },
    \end{equation}
    where $\ell_k(q_k)$ is defined in \eqref{eqn:blockdiagonalsubgauss:lk}-\eqref{eqn:blockdiagonalsubgauss:psi}.
    Then, the following bound holds 
    \begin{equation}
        \|\Bmat\| \leq \beta_{S_c},
    \end{equation}
    with probability at least 
    $\rho_{S_c}=1-\sum_{k=1}^K 2e^{-q_k}$. 
\end{theorem}

%% file: numerical_results.tex
\section{Numerical Results}\label{sec:numerical}
We now illustrate the behaviour of the generalization error with data from the distributions discussed in the preceding sections,
as well as image data from the MNIST dataset \cite{lecun-mnisthandwrittendigit-2010}.

We first explain the experimental setup for with the synthetic datasets.
We consider the following distributions for $\avec$:
a) Isotropic Gaussian (Iso.~G.) with  $\Nc(0,\eye{p})$;
    b) Correlated Gaussian (Corr.~G.) with $\Nc(0,\bar{\Sigmamat})$ with a  non-diagonal $\bar{\Sigmamat}\in\Rbb^{p\times p}$;
    c) Bernoulli (Bern.) distribution on $\{-1,1\}$ with $\Sigmamat=\eye{p}$, i.e., $a_{ij}$ is $-1$ or $1$  with probability $1/2$. 
These constitute examples for the settings of Section~\ref{sec:gaussian:iid}, Section~\ref{sec:gaussian:correlated} and Section~\ref{sec:subgaussian}, respectively. 
As our example for the sub-gaussian distributions, 
we consider the Bernoulli distribution
which is commonly used, for example, in compressive sensing literature \cite{foucartRauhut_2013}.
The covariance matrix $\bar{\Sigmamat}=\Umat\Lambdamat\Umat\T$ is fixed throughout the experiments and chosen as follows: $\Umat\in\Rbb^{p\times p}$ is sampled from a Haar distribution \cite{zhao_distributed_2015} and the eigenvalues are given by $\Lambdamat =\diag(\mu_i)\in\Rbb^{p\times p}$, 
with $\tilde{\mu}_{i+1} = 0.9631\tilde{\mu}_i$ and $\mu_i = p\tilde{\mu}_i/\sum_{i=0}^{p-1} \tilde{\mu}_i$. 
The parameter vector $\xvec$ is fixed for all experiments,
randomly chosen with i.i.d. uniform elements on $[-1, 1]$ and normalized so that $\|\xvec\|=1$.
We set $n=75$,  $p=200$ and use a network of $K=2$ nodes, hence $p = p_1+p_2$. Algorithm~\ref{alg:cocoa} is run for $T=1000$ iterations with $\lambda=0$ unless otherwise stated. 
The generalization error is reported as the emprical mean-squared error (MSE) which is calculated as
    $
        \text{MSE} \triangleq
        \frac{1}{{\bar{n}}N}
        \sum_{i=1}^N
        \left\|
            \Amat_{test,(i)} (\xvec - \xhat_{(i)}^T))
        \right\|^2\!. 
    $
Average simulation results for $N = 100$ realizations of the training data $\Amat_{(i)},\,i=1,\,\dots,\,N$ are reported. 
Here, $\Amat_{test,(i)} \in\Rbb^{{\bar{n}} \times p}$ denotes the test data matrix for experiment $i$,
$\hat{\xvec}^T_{(i)}=\xhat_{(i)}^T(\Amat_{(i)})$ is the  solution found by Algorithm~\ref{alg:cocoa} after its final iteration $T$ under $\yvec_{(i)} =\Amat_{(i)} \xvec$, 
and ${\bar{n}}=100n$ is the number of observations (i.e. rows) in each $\Amat_{test,(i)}$. 
Unless otherwise stated, all plots provide the performance of the algorithm after convergence.

\begin{figure}[t]
    \centering
    \includegraphics{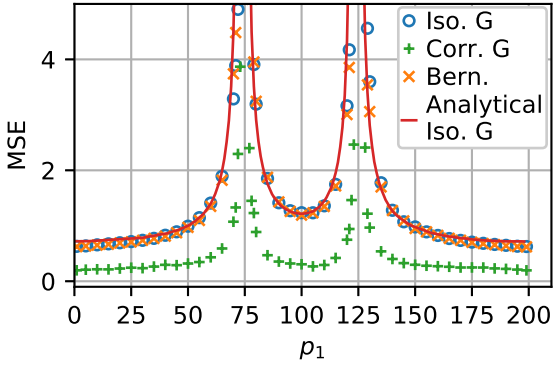}
    \caption{The generalization error for the three synthetic datasets together with the analytical expectation from Lemma \ref{lemma:conference_result}.}
    \label{fig:final_mse}
\end{figure}
\begin{figure}
    \centering
    \includegraphics{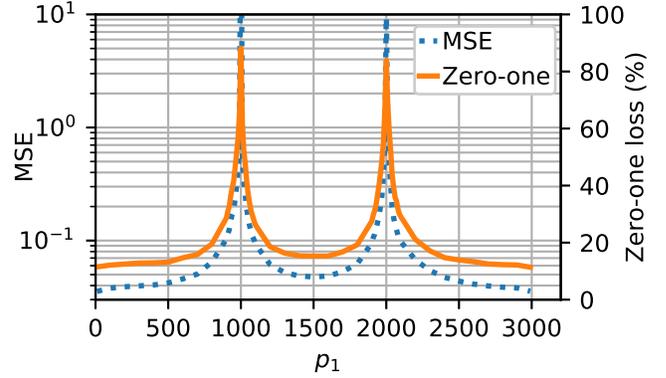}
    \caption{
        The generalization error in terms of MSE and zero-one loss for the MNIST example.
    }
    \label{fig:final_mse_mnist}
\end{figure}
\begin{figure*}[t]
    \centering
    \includegraphics[width=0.85 \linewidth]{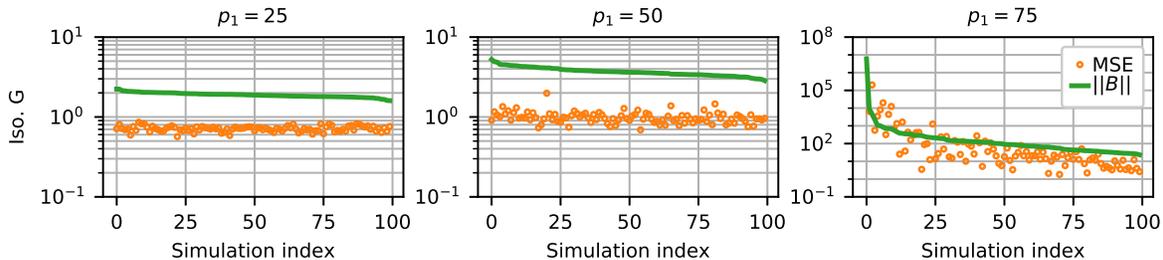} 
    \caption{
    Generalization error (MSE) versus $\|\Bmat\|$
    for each realization, i.e., simulation, of the training data.
    The simulation indices are ordered so that $\|\Bmat\|$ decreases monotonically with increasing simulation indices.
    }
    \label{fig:svd_grid}
\end{figure*}

In addition to the above, we also consider the digit classification problem from the MNIST dataset \cite{lecun-mnisthandwrittendigit-2010,garcia_2018} in order to further illustrate the practical implications of our results.   
This dataset poses a classification problem consisting of ten classes, 
i.e., digits. 
We convert each  $28$-by-$28$  image  to a $784$-by-$1$ vector $\zvec_i$ and transform the data using the following random features \cite{RahimiRecht_2008}:
    $\avec_i = [\cos(\zvec_i\T \omegavec_1); ...; \cos(\zvec_i\T \omegavec_p)]\in\Rbb^{p\times 1}$, $p=3\times 10^3$
where $\omegavec_i\sim\Nc(0, \zeta^2 \eye{784})$ with $\zeta=0.2$\:. 
The matrix of regressors $\Amat\in\Rbb^{n\times p}$ is obtained by using $\avec_i\T$ as its rows.
We train one classifier for each class and apply a one-v.s.-rest classification strategy \cite{bishop2006pattern}.
We subsample the training dataset with a factor of $60$, resulting in $n=10^3$ samples. For the test, we use the full test dataset with ${\bar{n}}=10^4$.
We report both the MSE and the classification error on the test data. 

 \kern-0.5em
\subsection{Generalization Error and the Partitioning of the Model}\label{sec:numerical_a}

In Figure \ref{fig:final_mse}, 
we present the empirical generalization error associated with the solution of \cocoa{} (Algorithm~\ref{alg:cocoa}) as $p_1$ is varied from $1$ to $p-1$.
The results for the three synthetic datasets together with the theoretical expected generalization error from  Lemma~\ref{lemma:conference_result} for the isotropic Gaussian case (Analytical~Iso.~G) are provided.
These plots illustrate that the generalization error depends significantly on the partitioning.
For all datasets, 
the average generalization error blows up as either $p_1$ or $p_2$ approaches $n$,
and it is relatively low when $p_1$ and $p_2$ are both far from $n$.
In particular, the peak MSEs are given by $2.2\times 10^9$, $8.1\times 10^2$ and $1.2\times 10^6$ for the cases a) -- c), respectively. 
Note that these values are comparably large and far outside the range of the plot, 
hence they are truncated, 
in Figure~\ref{fig:final_mse}. 

These observations are consistent with
Theorems~\ref{theorem:Bbound_prob} -- \ref{thm:subg_2},
demonstrating that the generalization error is small with high probability when $p_1$ and $p_2$ are far from $n$,
while small values cannot be guaranteed when $p_1$ or $p_2$ are close to $n$.
We now report the performance of the centralized solution in \eqref{eqn:centralized_sol}.
For all the cases, i.e.,  Iso. G, Corr. G and Bern.,
the training error is below $10^{-21}$ for the centralized solution as well as for the distributed solution for all values of $p_1$ (values are not included in the plots).
The generalization error for the centralized solution is $0.63,~0.20$ and $0.62$ for the cases  Iso.G, Corr.G, Bern., respectively.

These results illustrate that the partitioning can greatly affect the generalization error, making it significantly larger than what the centralized solution achieves,
while the training performance is on the same level as the centralized solution. 
The plots for the Iso. G. data in Figure~\ref{fig:final_mse} illustrates a close match between the empirical average generalization error and the expectation results in Lemma~\ref{lemma:conference_result}.
This observation emphasizes that the result in Corollary~\ref{cor:pklarge:gaussiid:expectation} can be relevant even if $p_k\geq n$ is not fulfilled for all nodes.

\subsection{Generalization Error on MNIST data with \cocoa{}}
\label{sec:mnist_simulations}

In Figure~\ref{fig:final_mse_mnist}, we plot the MSE and the zero-one loss, i.e., the percentage of incorrect classifications, for the MNIST  test data. 
Similary as in Figure~\ref{fig:final_mse}, 
the generalization error significantly depends on the partitioning at the nodes, 
both in terms of MSE and zero-one loss. 
In particular, we see an extremely large error if any $p_k$ is close to $n$ compared to the case where $p_k$ and $n$ are significantly different.
In particular, the generalization error (in terms of MSE) with $p_k=n$ is given by $2.5\times 10^3$ whereas  the training error is below $10^{-27}$ for all choices of $p_1$.  
The centralized solution in \eqref{eqn:centralized_sol} also achieves a training MSE below $10^{-27}$ while the corresponding generalization error is $3.5\times 10^{-2}$.

These results highlight practical consequences of design choices in distributed learning. 
In particular, it is not only for data coming exactly from certain probability distributions, 
but also for practical real-world datasets that the generalization error significantly depends on the partitioning over the nodes. 
Furthermore, 
the results here together with the results for the synthetic data in Section~\ref{sec:numerical_a}
suggest that in order to have a low generalization error one should avoid a partitioning where $p_k$ is close to $n$ for any node.%

\kern-0.15em
\subsection{Generalization Error and Spectral Norm of $\Bmat$}
\kern-0.15em
Theorems~\ref{theorem:Bbound_prob}-\ref{thm:subg_2} highlight the dependence of the generalization error on the spectral norm of $\Bmat$.  
We now further investigate this relationship. For ease of disposition, we consider only the isotropic Gaussian data.
In Figure~\ref{fig:svd_grid}, we plot $\|\Bmat\|$ and the generalization error (the MSE) for each of the $100$ different realizations of the training dataset, i.e., $\Amat$, that we have averaged over in Section \ref{sec:numerical_a}.
Each simulation index corresponds to one realization of the training dataset, i.e., one realization of $\Amat$. 
For each simulation index, the corresponding spectral norm $\|\Bmat\|$ and the MSE is provided.
The simulation indices are arranged so that $\|\Bmat\|$ is monotonically decreasing from left to right. 

Comparing the plots for $p_1\in\{25,50,75\}$, we observe that the MSE level depends on $\|\Bmat\|$ in a consistent manner.
The partitioning $p_1=25$ gives the lowest values of $\|\Bmat\|$,
as well as the lowest values of the MSE.
When $p_1$ is increased to $p_1=50$, both $\|\Bmat\|$ and the MSE increases slightly.
Consistent with Remark~\ref{remark:isoG}, with $p_1=75$ (hence $p_1=n=75$, where $n$ is the number of observations)
both $\|\Bmat\|$ and the MSE start to take extremely large values, such as up to $10^5$ for the MSE.  

For $p_1=25$ and $p_1=50$, both $\|\Bmat\|$ and the MSE are concentrated around their mean over different simulation indices. This illustrates that with this type of partitioning, it is possible to obtain reliable performance over different training datasets. 
On the other hand, we observe an extremely large spread in the MSE (from $1$ to over $10^5$)  over different simulation indices when $p_1=75$,
illustrating how under this partitioning, 
the generalization performance can vary substantially over different realizations of the training dataset.

\begin{figure}[t]
    \centering
    \includegraphics[width=1\linewidth]{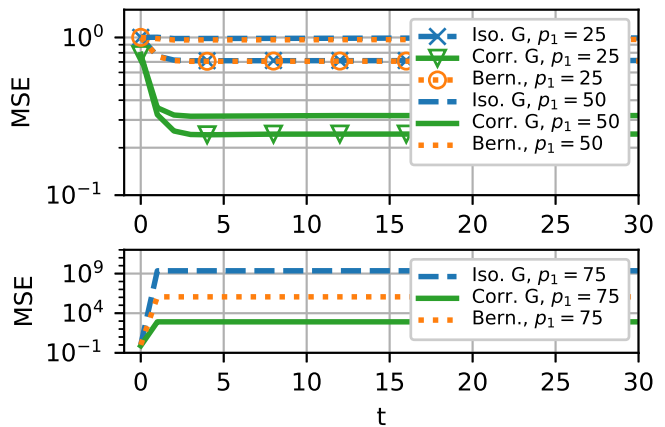}
    \caption{
        The average generalization error of the unregularized \cocoa{} evaluated at each iteration.
    }
    \label{fig:mse_time}
\end{figure}

\subsection{Behaviour of the Generalization Error over Iterations}
\label{sec:numerical_iterations}
We now investigate convergence of the generalization error over iterations of \cocoa{}. 
Furthermore, we verify the analytical result from Lemma~\ref{lemma:b_proj}.

In Figure~\ref{fig:mse_time}, the average generalization error associated with the solution produced by each iteration of \cocoa{} is plotted for the first $30$ iterations.
There is no visual change in the error values on the plot in the later iterations, hence this range is chosen to be able to better illustrate the transient behaviour. 

We observe that the algorithm on average converges  quickly, within the first few iterations for all cases.
The curves for $p_1=75$ are consistent with the result of Lemma~\ref{lemma:b_proj}: here all nodes have $p_k\geq n$, 
hence the algorithm converges in one iteration. 
(Although the plots shows only the average, this is also true for the individual runs.)
For the cases of $p_1=25$ and $p_1=50$,  although the results of Lemma~\ref{lemma:b_proj} does not directly apply,  the quick convergence suggests that  $\Bmat^t$ becomes an approximate projection matrix in the first iterations.

\begin{figure}[t]
    \centering
    \includegraphics[width=0.95 \linewidth]{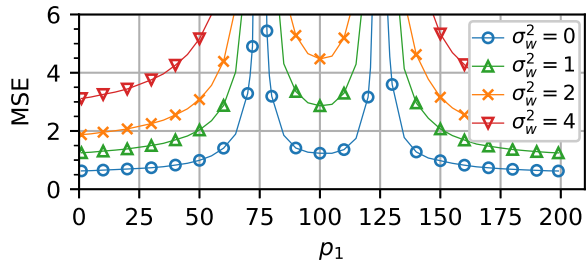}
    \caption{
        The average generalization error for the Iso. Gaussian case for  $\lambda=0$ and varying noise levels $\sigma_w^2$.
    }
    \label{fig:mse_noise}
\end{figure}
\begin{figure}[t]
    \centering
    \includegraphics[width=0.95 \linewidth]{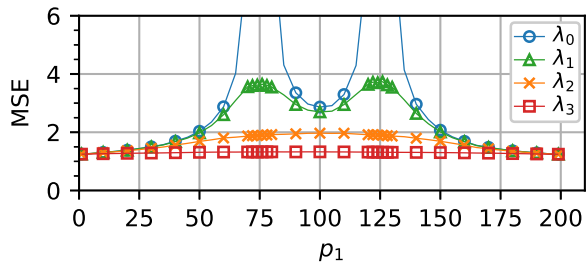}
    \caption{
        The average generalization error for the Iso. Gaussian case evaluated at $t\!=\!1000$ for varying values of $\lambda$
        Here, $(\lambda_0,\lambda_1,\lambda_2,\lambda_3)=(0,10^{-3},10^{-2},10^{-1})$,
        and $\sigma_w^2=1$.
    }
    \label{fig:mse_reg}
\end{figure}
\begin{figure}[t]
    \centering
    \includegraphics[width=0.92 \linewidth]{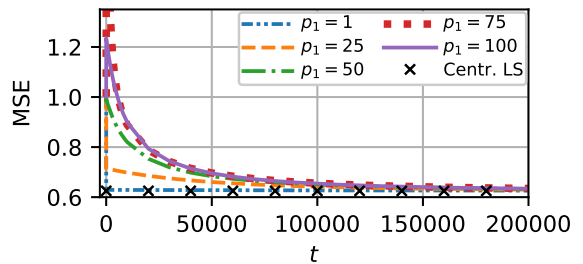}
    \caption{The average generalization error evaluated in each iteration for \cocoa{} with regularization $\lambda\!=\!10^{-3}$ and $\sigma_w^2\!=\!0$.}
    \label{fig:mse_reg_over_time}
\end{figure}

\subsection{Generalization Error and Noisy Training Data}\label{sec:numerical_noise}
We now focus on the effect of noise. In particular, we consider the setting with $\lambda=0$ with additive Gaussian noise in the training data,
i.e., with $w_i\sim\Nc(0,\sigma_w^2)$ in the model \eqref{eqn:model}.
Note that the test data is noise free,
and that the generalization error is defined as in \eqref{eqn:gen_err}. 
We consider the case with isotropic Gaussian regressors.  

In Figure~\ref{fig:mse_noise},  we present the generalization error for four different noise levels,
$\sigma_w^2=\{0, 1, 2, 4\}$.
We observe that the overall level of the generalization error increases with the noise level.
There are again large peaks in the error for $p_k$ values close to $n$,
suggesting that noisy data does not dampen the peaks,
further supporting the insights gained from our analytical results in Theorem~\ref{theorem:Bbound_prob} -- \ref{thm:subg_2}.
We emphasize that Figure~\ref{fig:mse_noise} shows the generalization error for $t=1000$, which is well after convergence. 
Our simulations show that the generalization error in the unregularized, noisy setting converges in the same quick fashion as the unregularized noise free setting as illustrated in Figure~\ref{fig:mse_time}.
Due to space limitations, these convergence plots are not included here.

\subsection{Generalization Error and Regularization}\label{sec:numerical_reg}
In the preceding sections, 
we have considered the unregularized scenario.
In this section, 
we provide results that illustrate that even with regularization
the partitioning scheme can have a large impact on the generalization error. 
In particular, the generalization error heavily depends on the partitioning before convergence.

We consider the scenario with 
${w_i\sim \Nc(0,1)}$ in \eqref{eqn:model} and with varying values of the regularization parameter $\lambda$ in \eqref{eqn:problem}.
In Figure~\ref{fig:mse_reg},
we compare the generalization error associated with the solutions produced by \cocoa{} at iteration $t=1000$,
with Iso.~G. regressors. 
We consider the following choices of $\lambda$: 
    $\lambda_0=0$, 
    $\lambda_1=10^{-3}$,
    $\lambda_2=10^{-2}$ and
    $\lambda_3=10^{-1}$.
The figure illustrates that the generalization error peaks are dampened for all three choices of $\lambda>0$,
compared to $\lambda=0$: 
For $\lambda_0=0$, the peak generalization error is $\approx 10^9$, whereas for $\lambda_1$ -- $\lambda_3$ the peak error values
are between $\approx 1$ and $\approx 4$.
The minimum generalization error over all partitioning choices is around $1$. 
Hence, 
with larger values of $\lambda$,
dependence on the partitioning can become weaker or completely vanish.
We note that, for $\lambda_1$, there is still a strong dependence on the partitioning,
and the error have significant peaks, although bounded, around $p_k\approx n$.
As we will discuss more in detail in the remainder of this section, convergence rate of \cocoa{} heavily depends on $\lambda$ and the algorithm has not yet converged for all values of $\lambda$ in Figure~\ref{fig:mse_reg}.

We now focus on the transient behaviour of the regularized \cocoa{} algorithm.
In Figure~\ref{fig:mse_reg_over_time}, 
we plot the average generalization error with $\lambda=10^{-3}$ and $\sigma_w^2=0$ 
over the iterations $t$
for five partitioning choices $p_1=\{1,25,50,75,100\}$ as well as that of the centralized regularized least-squares solution in \eqref{eqn:centralized_sol}.
We note two main effects:
Firstly,  for all choices of $p_1$, 
the generalization error converges to that of the centralized LS solution,
i.e., $\approx 0.63$.   
This is consistent with the fact that the regularized \cocoa{} 
converges to the centralized LS solution with the same regularization, 
see the discussions in Section~\ref{sec:gen_err}.  
Secondly, 
the partitioning can greatly affect the convergence rate of the regularized algorithm.
For values of $p_1$ closer to $n=75$,
the convergence rate is much slower than for values smaller than $p_1=25$.
For instance, 
for $p_1=25$ the MSE reaches $\approx 0.66$ at $t=5\cdot 10^{4}$,
while it takes almost twice as many iterations to reach the same MSE for $p_1=75$.

We conclude this discussion by highlighting the manner in which our results presented in
Theorem~\ref{theorem:Bbound_prob}--\ref{thm:subg_2}
are relevant for the regularized setting,
although they are derived for the unregularized version of \cocoa{}.
A key observation in Figure~\ref{fig:mse_reg_over_time} is that while the partitioning does not affect 
the performance after convergence,
it has a considerable effect on the conditioning of the problem; which in turn affects the practical performance of the algorithm significantly. 
If one would set $T$ in the order of $10^4$,
which is a very large number of iterations in terms of convergence in relation to the unregularized setting,
then the choice of partitioning can severely affect the generalization performance of the final solution $\xhat^T$, as seen in Figure~\ref{fig:mse_reg}.
In other words, 
if we are not willing or able to run the algorithm for a substantially larger number of iterations compared to the unregularized case, 
then we should also avoid having $p_k$ close to $n$ for the regularized setting,
in order to avoid a large generalization error,
similarly as we should avoid $p_k$ close to $n$ in the unregularized setting.

\subsection{Hyperparameter Study for \cocoa{} parameters } \label{sec:numerical_hyperparam}
\begin{figure}
    \centering
    \includegraphics{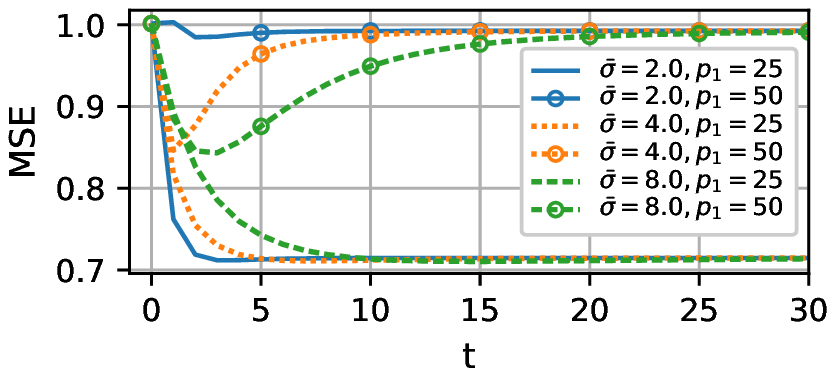}
    \caption{Convergence of \cocoa{} for three choices of $\subproblemp$.}
    \label{fig:mse_reg_hyperparams}
\end{figure}

We now conduct a hyperparameter study for the aggregation parameter $\aggregationp$ and the subproblem parameter $\subproblemp$ of \cocoa{}.
Choosing these parameters as
$\aggregationp\in (0,1]$ and $\subproblemp\geq \aggregationp K$ facilitates formal convergence guarantees  \cite[Sec. 3.1]{smith_cocoa_nodate}.
As discussed in Section~\ref{sec:gen_err},  with $\lambda=0$, 
it is only the ratio $\aggregationp/\subproblemp$ that affects our expressions, hence our previous results cover all admissible values of  $\aggregationp$ and $\subproblemp$   as long as  $\aggregationp/\subproblemp =K$. 
To study scenarios that have not been covered with the previous plots but still with formal convergence guarantees, we use $\aggregationp=1$ and $\subproblemp\in\{ 4, 8\}$ together with the earlier case of $\subproblemp =K=2$ for comparison. 
In Figure~\ref{fig:mse_reg_hyperparams}, we present the convergence under isotropic Gaussian setting,
with $\lambda=0$, $\wvec=0$ and $p_1 \in \{25, 50\}$.
While the error converges to the same value for all choices of $\subproblemp$,
the rate of convergence is slower for $\subproblemp$ with $ \subproblemp> K \aggregationp$,
as expected. 
Hence, these results support the expectation that the parameters chosen for the previous studies in this section provide relatively fast convergence. 

%% file: discussion.tex
\section{Discussions}\label{sec:discussions}
We now discuss the practical guidelines our results provide. 
Our results emphasize the relation between the partitioning of the model unknowns over the network and the generalization error.
Although the training error is low and at the same level as that of the centralized solution,
a low generalization error is not guaranteed when the number of unknowns $p_k$   at any node  is close to the number of observations $n$ in the training data.
Furthermore, if any $p_k$ is close to $n$,
then the generalization performance can vary significantly over different realizations of the training data.  
Hence, the partitioning should not be chosen so that $p_k$ is close to $n$, for any node.

Explicit regularization,
i.e., $\lambda>0$ in \eqref{eqn:problem},
can improve the generalization error of the distributed scheme. 
Nevertheless, 
the choice of the regularization parameter $\lambda$ is not straightforward.
The algorithm can need significantly different numbers of iterations (such as $10^4$ times more) for different values of $\lambda$.
If $\lambda$ is chosen too small,
then the generalization error can still be relatively large if any $p_k$ is close to $n$, 
compared to other possible data partitionings.
Hence, for a fixed number of iterations,
one should choose a large enough $\lambda$ in order to guarantee that the generalization error does not depend on the data partitioning over nodes.

While our results are restricted to the convex setting,
extensions into the non-convex formulations are considered an important line of future research.
A key step for this challenging set-up could involve relaxation of the definition of convergence, as done in the centralized case \cite{raginsky2017nonconvex, Gelfand1991RecursiveSA}.

%% file: conclusions.tex
\kern-0.25em
\section{Conclusions}\label{sec:conclusions}
We have focused  on the generalization error associated with solutions produced by the distributed learning algorithm \cocoa{} for the linear regression problem.
We have presented upper bounds on the generalization error that hold with high probability for  isotropic Gaussian, correlated Gaussian and sub-gaussian data.
We have compared our probabilistic bounds with the results on the expected generalization error.  With our numerical results, we have illustrated the generalization performance of the algorithm with both synthetic and real data.

In existing works, there is a lack of efforts for determining how the generalization error in the distributed setting can be affected by the algorithm design. 
Here, we have addressed this gap by providing bounds
that characterize how the partitioning of the model's unknowns over the nodes in the network affects the generalization error.
Our results provide  guidelines on how to partition the model over the network in order to avoid potential pitfalls. %
Our results show that if the number of unknowns $p_k$ in any node  is close to the total number of observations $n$, then the generalization error can be very large,
even though the training error is small.
Hence, in order to obtain a good generalization performance, the number of unknowns in any node should be chosen to be sufficiently larger or smaller than the number of observations if possible.  
If one has to operate with a node with $p_k\approx n$, regularization can be used to significantly dampen the generalization error. On the other hand, 
choosing the regularization parameter is not straightforward. 
If the regularization parameter is too small, 
\cocoa{} needs a relatively large amount of iterations in order to mitigate the effect that the partitioning has on the generalization error.

Extensions of our results to the fully decentralized scenarios as well model misspecification are considered as important directions for future work.

%% file: appendix/appendix.tex
\kern-0.25em
\section{Appendix}
\kern-0.25em
\setlength{\abovedisplayskip}{2pt}
\setlength{\belowdisplayskip}{2pt}

\input{appendix/preliminaries}
\kern-1em
\subsection{Proof of Lemma~\ref{lemma:b_proj}}\label{sec:lemma:b_proj}
    Expanding $\Bmat^2$, we obtain the following 
    \begin{equation}
        \Bmat^2 
        = (\eye{p} - \frac{1}{K}\Abar\Amat)^2 
        = \eye{p} + \frac{1}{K^2}(\Abar\Amat)^2 - \frac{2}{K}\Abar\Amat,
    \end{equation}
    Note that $(\Abar \Amat)^2 = \Abar \Amat \Abar \Amat$,
    where $\Amat\Abar = \sum_{k=1}^K \Amat_k\Amat_k\p$.
    The rows of $\Amat_k$ have positive definite covariance matrices.
    Hence, by Property~\ref{prop:gauss_identity} under $n\leq p_k$, we have
    $\Amat_k\Amat_k\p = \eye{n},$
    and $\Amat\Abar = K\eye{n}$.
    Hence, $(\Abar\Amat)^2 = K \Abar\Amat$. Then, we have 
    \begin{equation}
        \Bmat^2 = \eye{p} + \frac{1}{K^2}K\Abar\Amat - \frac{2}{K}\Abar\Amat
        = \eye{p} - \frac{1}{K}\Abar\Amat = \Bmat.
    \end{equation}  

\kern-1em
\subsection{Proof of Theorem \ref{theorem:Bbound_prob}} \label{proof:theorem_Bbound_prob}

    Using the triangle inequality and \eqref{eqn:xhattp:B},
    we obtain
    \begin{equation}\label{eqn:B_Bound}
        \|\Bmat\| \leq 
        1 + \frac{1}{K}\left\|\Abar\Amat\right\|. 
    \end{equation}
    We now present the following algebraic property of $\Abar \Amat$ which holds regardless of the distribution of $\Amat$:
    \input{props/lemma_Abar_A_bound}
    
    \noindent Proof: See Section~\ref{proof:lemma:Abar_A_bound}.
    Note that we in the subsequent sections use that 
    $\frac{1}{\sigma_{min+}}\leq\frac{1}{\sigmamin}$,
    but when $\sigmamin\rightarrow 0$, this upper bound is uninformative.
    The aim of our results is to find bounds on the minimum singular value which is away from zero.

    The result of Theorem~\ref{theorem:Bbound_prob} is obtained by combining \eqref{eqn:B_Bound}, \eqref{eqn:Abar_A_bound_upper}  and the bounds on the extreme singular values of $\Amat_k$. 
    In particular, denote the event that the singular value inequalities given in Lemma~\ref{lemma:tracy_widoms_wisdom} holds for the partition $\Amat_k$ as $c_k$, i.e. $c_k = c_k^L \cap c_k^U$, where $c_k^L = \{ r_k^L(q_k)
            \leq \sigma_{\min}(\Amat_k)\}$ and $c_k^U =\{ \sigma_{\max}(\Amat_k) \leq r_k^U(q_k) \}$,  
    with $r_k^L(q_k)=  \sqrt{\rmaxk} - \sqrt{\rmink} - q_k$ and  $r_k^U(q_k)=  \sqrt{\rmaxk} + \sqrt{\rmink} + q_k$.  
    Note that $c_k$ can be rearranged as follows
    \begin{align}\label{eqn:ck:Aiid:reverse}
      c_k \!=\!  \Big\{ 
        \frac{1}{\sigma_{\min}(\Amat_k)} 
        \leq \frac{1}{r_k^L(q_k)} \cap 
        \sigma_{\max}(\Amat_k) 
        \leq r_k^U(q_k) \Big\}.
    \end{align}

    For any  $k\neq i$,  $c_k$ and $c_i$ are statistically independent since the entries of $\Amat$ are statistically independent.
    Hence,
    \begin{align}\label{eqn:ck:Aiid:prob}
        \Pr\left({ \bigcap_{k=1}^K} c_k\right) 
        \geq {\prod_{k=1}^K} (1-2e^{-q_k^2/2})_+
    \end{align}
    
    Therefore, \eqref{eqn:Abar_A_bound_upper}, \eqref{eqn:ck:Aiid:reverse} and  \eqref{eqn:ck:Aiid:prob}  yields to 
    \begin{equation}\label{eqn:Abar_A_upper_prob}
    \| \Abar \Amat \|^2 
      \leq  K +   \sum_{k=1}^K  \sum_{\substack{i=1\\i\neq k}}^K 
    \tfrac{
        (\sqrt{r_{\max,k}}  +  \sqrt{r_{\min,k}} + q_k)^2
    }
    {
        (\sqrt{r_{\max,i}}  -  \sqrt{r_{\min,i}} - q_i)^2    
    },
\end{equation} 
with probability at least $\prod_{k=1}^K(1-2e^{-q_k^2/2})_+^K$. The desired result in Theorem~\ref{theorem:Bbound_prob} is obtained by combining \eqref{eqn:Abar_A_upper_prob} and \eqref{eqn:B_Bound}. 
We bound $q_i < \sqrt{r_{\max,i}}  -  \sqrt{r_{\min,i}}$, so that the bound on $\sigmamin^2(\Amat_i)$ is informative, i.e., strictly greater than zero.
    
\kern-1em
\subsection{Proof of Lemma \ref{lemma:Abar_A_bound}}
\kern-0.25em
\label{proof:lemma:Abar_A_bound}
    
    The matrices $\Abar\Amat$ and $\Amat\T\Abar\T\Abar\Amat$ can be seen as matrices consisting of $K\times K$ blocks.
    The $(k,j)$\th{} block of $\Abar\Amat$ is of size $p_k\times p_j$ and given by
    \begin{equation}\label{eqn:blocks_abar_a}
        [\Abar\Amat]_{k,j} = \Amat_k\p\Amat_j.
    \end{equation}
    The $(k,j)$\th{} block of $\Amat\T\Abar\T\Abar\Amat$ is of size $p_k\times p_j$ and given by
    \begin{equation}\label{eqn:blocks_aT_abarT_abar_a}
        [\Amat\T\Abar\T\Abar\Amat]_{k,j} 
        = \Amat_k\T\left(\sum_{i=1}^K {\Amat_i\p}\T\Amat_i\p\right)\Amat_j.
    \end{equation}
    
    Hence, we have 
    \begin{align}
        \|\Abar\Amat\|^2 \!=\! \|\Amat\T\Abar\T\Abar\Amat\| \! &\leq \!
        \sum_{k=1}^K\| \Amat_k\T (\sum_{i=1}^K {\Amat_i\p}\T \Amat_i\p)\Amat_k \|\label{eqn:Abar_A_bound:normdecomposition}\\
         &\leq \sum_{k=1}^K\sum_{i=1}^K 
        \| \Amat_k\T {\Amat_i\p}\T \Amat_i\p\Amat_k \|,\label{eqn:Abar_A_bound:triangle}
    \end{align}
    where we obtained \eqref{eqn:Abar_A_bound:normdecomposition} using property \ref{hölders_trick} of Section \ref{sec:preliminaries} repeatedly on the blocks of  $\Amat\T\Abar\T\Abar\Amat$ on the diagonal (i.e. $k=j$). In \eqref{eqn:Abar_A_bound:triangle}, we used the triangle inequality.

    We now consider the individual terms in the double summation of \eqref{eqn:Abar_A_bound:triangle}. 
    For the terms with $k\!=\!i$, consider the s.v.d. $\Amat_k = \Umat_k \Lambdamat_k \Vmat_k\T$, where
    $\Lambdamat_k\in\Rbb^{n \times p_k}$ is the (possibly rectangular) diagonal matrix of singular values, and $\Umat_k\in\Rbb^{n \times n}$  and $\Vmat_k \in\Rbb^{p_k \times p_k}$ are unitary. 
    Hence, we have
    $
        \Amat_k\T {\Amat_k\p}\T \Amat_k\p \Amat_k = 
          \Vmat_k \Lambdamat_k\T {\Lambdamat_k\p}\T \Lambdamat_k^+  \Lambdamat_k \Vmat_k\T = \Vmat_k \Rmat_k \Vmat_k\T,
    $
    where $\Rmat_k\in\Rbb^{p_k\times p_k}$ is a diagonal matrix with ones and zeroes on its diagonal. 
    Hence, 
    \begin{equation}\label{eqn:AmatkAmatkp:norm}
        \|\Amat_k\T {\Amat_k\p}\T \Amat_k\p \Amat_k\| = 1.
    \end{equation}
    For the terms with $i\neq k$, we use that the spectral norm 
    is submultiplicative and self-adjoint \cite[Sec. 5.6]{horn1990matrix} to obtain
    \begin{equation}\label{eqn:AmatkAmatip:norm}
        \|\Amat_k\T {\Amat_i\p}\T \Amat_i\p\Amat_k \| \leq \|\Amat_k\|^2\|\Amat_i\p\|^2 = \frac{\sigma_{\max}^2(\Amat_k)}{\sigma_{\min+}^2(\Amat_i)}
    \end{equation}
    where we have used $\|\Amat_k\|^2 =\sigma_{\max}^2(\Amat_k)$  and the property 
    $\|\Amat_i\p\| = \frac{1}{\sigma_{\min+}(\Amat_i)}$ where $\sigma_{\min+}(\Amat_i)$ is the smallest non-zero singular value of $\Amat_i$.  
    Note that $\Amat_i\p$ can be written as $\Amat_i\p = \Vmat_i \Lambdamat_i\p \Umat_i\T$%
    .
    Hence, $\|\Amat_i\p\| = \sigma_{\min+}^{-1}(\Amat_i)$. 
    Now combining \eqref{eqn:AmatkAmatkp:norm},  \eqref{eqn:AmatkAmatip:norm} and  \eqref{eqn:Abar_A_bound:triangle}, we obtain \eqref{eqn:Abar_A_bound_upper}.

\kern-1em
\subsection{Proof of Lemma \ref{lemma:alternative_bound}}
    \label{proof:lemma_alt_bound}
    Let $\Mmat = \Bmat\T\Bmat\in\Rbb^{p\times p}$, and denote its $K$ blocks on the diagonal as $\Mmat_{kk}\in\Rbb^{p_k\times p_k}$, $\forall k$.
    Using Property~\ref{hölders_trick} of Section~\ref{sec:preliminaries}, %
    \setlength{\abovedisplayskip}{0pt}%
    \setlength{\belowdisplayskip}{0pt}%
    \begin{align}\label{eqn:error1:Mkk}
        \|\Mmat\| \leq {\sum_{k=1}^K} \|\Mmat_{kk}\|.
    \end{align}
    \setlength{\abovedisplayskip}{2pt}%
    \setlength{\belowdisplayskip}{2pt}%
    With $\Bmat$ from \eqref{eqn:xhattp:B},
    we have $\Mmat=\Bmat\T\Bmat=\eye{p} + \frac{1}{K^2}\Amat\T\Abar\T\Abar\Amat - \frac{1}{K}(\Abar\Amat)\T - \frac{1}{K}\Abar\Amat$.
    Using \eqref{eqn:blocks_abar_a} and \eqref{eqn:blocks_aT_abarT_abar_a},
    we find that the blocks on the diagonal of $\Mmat$, 
    $\Mmat_{kk}$ can be decomposed such that 
    $\Mmat_{kk} = \Cmat_{kk} + \Dmat_{kk}$, 
    where $\Cmat_{kk} = (\eye{p_k} - \tfrac{1}{K}\Amat_k\p \Amat_k)^2$
        and 
    $\Dmat_{kk} 
        = \frac{1}{K^2}\Amat_k\T(
            \sum_{\substack{i=1\\i\neq k}}^K {\Amat_i\p}\T\Amat_i\p
            )\Amat_k$.
            
    By Property~\ref{prop:gauss_identity},
    under $n\geq p_k$,
    we have $\Amat_k\p\Amat_k=\eye{p_k}$, 
    hence
    $\Cmat_{kk} = \eye{p_k} (\frac{K-1}{K})^2$ and 
    $\|\Cmat_{kk}\| = (\frac{K-1}{K})^2$.
    From the proof of Theorem~\ref{theorem:Bbound_prob}, 
    we have with probability at least 
    $\Bar{\rho}_G = \prod_{k=1}^K(1-2e^{-q_k^2/2})_+$:
    \begin{equation}
        \|\Dmat_{kk}\|=\frac{1}{K^2}
        \|\Amat_k\T (
            \sum_{\substack{i=1\\i\neq k}}^K
            {\Amat_i\p}\T\Amat_i\p
        )\Amat_k\| 
        \leq \frac{1}{K^2} 
        \sum_{\substack{i=1\\i\neq k}}^K\Bar{\gamma}_{k,i},
    \end{equation}
    for $\krange$,
    with 
    $\Bar{\gamma}_{k,i} 
        = \frac{(\sqrt{\rmaxk} + \sqrt{\rmink} + q_k)^2}
            {(\sqrt{\rmaxi}-\sqrt{\rmini} - q_i)^2}.$
    Hence, using \eqref{eqn:error1:Mkk} and  $\|\Mmat_{kk}\|= \|\Cmat_{kk} + \Dmat_{kk}\| \leq   \|\Cmat_{kk}\| + \|\Dmat_{kk}\| $, we have
    \begin{align}
        \|\Bmat\T\Bmat\|
        \leq \frac{(K-1)^2}{K} + 
            \frac{1}{K^2} \sum_{k=1}^K\sum_{\substack{i=1\\i\neq k}}^K
            \Bar{\gamma}_{k,i}.
    \end{align}
    To conclude the proof, use the fact that
    $\|\Bmat\| = \sqrt{\|\Bmat\T\Bmat\|}$.

\kern-1em

\subsection{Proof of Lemma \ref{lemma:conference_result}}\label{sec:proof:conference_result}
    By \eqref{eqn:noisy_egen_Bbound:termstogether} and $\Sigmamat=\eye{p}$, we have 
    \begin{align}
        \egen(\xhat^1) 
        & = \|\Bmat \xvec - \frac{1}{K}\Abar \wvec\|^2 \\
        & = \|\Bmat \xvec \|^2 
        + \|\frac{1}{K} \Abar \wvec \|^2
        -\frac{2}{K} \xvec\T \Bmat\T \Abar \wvec.
    \end{align}
    Since $\wvec$ is zero-mean and statistically independent with $\Amat$, we have 
    \begin{align}\label{eqn:proof_sum_expected_norms}
        \Ebb_{\Amat,\wvec}[\egen(\xhat^1)]
        & = \Ebb_{\Amat}[\|\Bmat \xvec\|^2]
        + \frac{1}{K^2}\Ebb_{\Amat,\wvec}[\|\Abar\wvec\|^2].
    \end{align}
    The first term is evaluated in \cite[Thm.~1]{HellkvistOzcelikkaleAhlen_distributed2020_spawc}.
    We now focus on the second term.
    Using the fact that $\Amat$ and $\wvec$ are statistically independent,
    we find that
    \begin{align}
        \Ebb_{\Amat,\wvec}\big[\|\Abar\wvec\|^2\big]
        & = \Ebb_{\wvec}\big[ \wvec\T
            \Ebb_{\Amat}\big[
                \Abar\T\Abar
            \big]
            \wvec
        \big],
    \end{align}
    where $\Abar\T\Abar=\sum_{k=1}^K (\Amat_k\Amat_k\T)\p$.
    From \cite{cook_mean_2011} we have that
    $
    \Ebb_{\Amat}\big[ \Abar\T\Abar \big]
    = \frac{1}{n} \sum_{k=1}^K \gamma_k \eye{n},
    $
    with $\gamma_k$ as in \eqref{eqn:gamma_main}.
    Hence,
    \begin{equation}
        \Ebb_{\Amat,\wvec}\big[ \|\Abar\wvec\|^2 \big]
        \! = \! \frac{1}{n}\sum_{k=1}^K\gamma_k \Ebb_{\wvec}[\wvec\T\wvec]
        \! = \! \frac{1}{n}\sum_{k=1}^K\gamma_k n\sigma_w^2.
        \label{eqn:errorwithnoise}
    \end{equation}
    Combining \eqref{eqn:errorwithnoise}, \cite[Thm.~1]{HellkvistOzcelikkaleAhlen_distributed2020_spawc} and \eqref{eqn:proof_sum_expected_norms} concludes the proof.
\subsection{Proof of Theorem \ref{theorem_correlated_Bbound_prob}}
\kern-0.25em
    We first consider an intermediate general result on sub-gaussian variables. Background information on sub-gaussian variables and the notation can be found in Section~\ref{sec:preliminaries:subgaussian}. 

    \label{proof:theorem_correlated_Bbound_prob}
    \begin{lemma}\label{lemma:subg_tracy_widom}
        Let $\Mmat\in\Rbb^{n \times p}$ have rows i.i.d. with $\Sc(\Sigmamat)$, $\Sigmamat\in\Rbb^{p\times p}$.
        The following bounds hold with probability at least $1-2e^{-q}$, $q\geq 0$:
        \begin{align}\begin{split}\label{eqn:subg_tracy_widom_tall}
            &n\sigmamin(\Sigmamat) - \ell(q) \leq \sigmamin^2(\Mmat)  \\
            &\leq \sigmamax^2(\Mmat) \leq n\sigmamax(\Sigmamat) + \ell(q),
        \end{split}\end{align} 
        where 
        \begin{equation}\label{eqn:lq:def:subgaussian}
            \ell(q) =  C L^2 \left( \sqrt{\frac{p+q}{n}} + \frac{p+q}{n} \right)n\sigmamax(\Sigmamat),
        \end{equation}
        where $C$ is an absolute constant,
        and $L\geq 1$ is constant such that 
        $
          \psi_{\mvec}(\Sigmamat,\hvec)
            \leq L\, \sqrt{\hvec\T \Sigmamat \hvec},\,\forall \hvec\in\Rbb^{p\times 1}
        $ where $\mvec$ comes from the same distribution as the rows of $\Mmat$.
    \end{lemma}  
        
    \noindent Proof: See Section~\ref{proof:lemma:subg_tracy_widom}.

    In particular,  we have the following for the Gaussian case: 
    \begin{lemma}\label{lemma:Gaussian:singularValues}
        Let $\Mmat\in\Rbb^{n \times p}$ have rows i.i.d. with $\Nc(0, \Sigmamat)$, 
        $\Sigmamat\in\Rbb^{p\times p}$.
        Then \eqref{eqn:subg_tracy_widom_tall} holds with $L=\sqrt{8/3}$ in \eqref{eqn:lq:def:subgaussian},
        with probability at least $1-2e^{-q},\,q\geq 0$. 
        Additionally, the following holds with probability at least $1-2e^{-\Bar{q}^2/2}$, $\Bar{q}\geq0$:
        \begin{equation}\label{eqn:correlated_gauss_tw_broad}
            \sigmamin(\Mmat)
            \geq \sqrt{\sigmamin(\Sigmamat)}
            (\sqrt{p} - \sqrt{n} - \Bar{q}).
        \end{equation}
        
    \end{lemma}
    \noindent Proof: See Section~\ref{sec:proof:lemma:Gaussian:singularValues}.

    To prove Theorem \ref{theorem_correlated_Bbound_prob},
    we apply Lemma~\ref{lemma:Abar_A_bound}
    and Lemma~\ref{lemma:Gaussian:singularValues} to
    \eqref{eqn:B_Bound}.
    Note that the rows of $\Amat_k$ are i.i.d. with $\Nc(0,\Sigmamat_k)$,
    where $\Sigmamat_k$ is the $k$\th{} principal submatrix of $\Sigmamat$.
    Applying \eqref{eqn:subg_tracy_widom_tall} for  $\Amat_k$, we have \begin{align}\begin{split}\label{eqn:corr_gauss_tw_tall_Ak}
        &n\sigmamin(\Sigmamat_k) - \ell_k(q_k) \leq \sigmamin^2(\Amat_k) \\
        &\leq \sigmamax^2(\Amat_k) \leq n\sigmamax(\Sigmamat_k) + \ell_k(q_k),
    \end{split}\end{align}
    with probability at least $1-2e^{-q_k}$, $q_k\geq 0$,
    where 
    $\ell_k(q_k) = \tfrac{8}{3}
    n\sigmamax(\Sigmamat_k)C(\sqrt{\tfrac{p_k+q_k}{n}} + \tfrac{p_k+q_k}{n})$. 
    Additionally, by \eqref{eqn:correlated_gauss_tw_broad} the following holds with probability at least $1-2e^{-\Bar{q}_k^2/2}$, $\Bar{q}_k\geq 0$
    \begin{equation}\label{eqn:corr_gauss_tw_broad_Ak}
        \sigmamin(\Amat_k) \geq \sqrt{\sigmamin(\Sigmamat_k)}(\sqrt{p_k} - \sqrt{n} - \Bar{q}_k).
    \end{equation}
    For the lower bounds in Theorem \ref{theorem_correlated_Bbound_prob}, 
    we use
    \eqref{eqn:corr_gauss_tw_broad_Ak} for all broad matrices
    $\Amat_k$. 
    For tall matrices $\Amat_k$, \eqref{eqn:corr_gauss_tw_broad_Ak} is uninformative, then we use \eqref{eqn:corr_gauss_tw_tall_Ak}.
    For \eqref{eqn:thm_correlated_Bbound_prob_upper},
    we use \eqref{eqn:corr_gauss_tw_tall_Ak} for all matrices $\Amat_k$, $\forall k$,
    because we use the upper bound on the largest singular value for all $k$.
    Consider $\Kc = \{k:n<p_k\}$ which denotes the set of partition indices $k$ for which the matrices $\Amat_k$ are broad.
    Then, 
    by property \ref{sec:prob_intersect} in Section \ref{sec:preliminaries}, 
    the bounds in \eqref{eqn:corr_gauss_tw_tall_Ak} for all $\Amat_k$ and the bound in \eqref{eqn:correlated_gauss_tw_broad} for $k\in\Kc$ simultaneously hold with probability at least
    $
        1 - \sum_{k=1}^K 2e^{-q_k} - \sum_{k\in\Kc} 2 e^{-\Bar{q}_k^2/2},
    $
    which concludes the proof.
    
\kern-1em
\subsection{Proof of Lemma~\ref{lemma:subg_tracy_widom}}
\label{proof:lemma:subg_tracy_widom}
\kern-0.25em
    We use the following result from \cite{vershynin2018high}:
    \begin{lemma}{\emph{\cite[Sec. 4.7]{vershynin2018high}}}
        \label{lemma:tail_bound}  
        Consider the setting of Lemma~\ref{lemma:subg_tracy_widom}.
        Then, the following bound holds with probability at least $1-2e^{-q}$, $q \geq 0$:
        \begin{equation}\label{eqn:tail_bound}
            \|\Mmat\T\Mmat - n\Sigmamat\| 
            \leq \ell(q),
        \end{equation}
        where $\ell(q)$ is defined as in Lemma~\ref{lemma:subg_tracy_widom}.
        
    \end{lemma}
    Using Property~\ref{sec:prop_spectral_norm_bounds} from Section~\ref{sec:preliminaries}, we observe that if
    $
        \| \Mmat\T\Mmat - n\Sigmamat\| \leq \ell(q),
    $
    then
    $
        \sigmamin^2(\Mmat) \geq \sigma_{\min}(\Mmat\T\Mmat) \geq n \sigmamin(\Sigmamat) - \ell(q),
    $
    which constitutes the lower bound of Lemma~\ref{lemma:subg_tracy_widom}.
    To find the upper bound in \eqref{eqn:subg_tracy_widom_tall}, we
    apply the reverse triangle inequality to \eqref{eqn:tail_bound} to obtain
    $
        \ell(q) \geq \|\Mmat\T\Mmat\| - n\sigmamax(\Sigmamat),
    $  
    and use that
    $
        \sigmamax^2(\Mmat) = \|\Mmat\T\Mmat\|.
    $
    The upper and lower bounds hold with the same probability as in \eqref{eqn:tail_bound}. 

\kern-1em
\subsection{Proof of Lemma~\ref{lemma:Gaussian:singularValues}}
\label{sec:proof:lemma:Gaussian:singularValues}
\kern-0.25em
   
    Using \eqref{eqn:def:subgaussiannorm:singlevariable}, \eqref{eqn:def:subgaussvectornorm},
    the properties of Gaussian integral and the fact that  the rows  $\Mmat$ are i.i.d. with  $\Nc(0,\Sigmamat)$,  we arrive at $ \psi_{\mvec}(\Sigmamat, \hvec)= \sqrt{\frac{8}{3}}\sqrt{\hvec\T\Sigmamat\hvec} $, hence $L$ can be chosen as $\sqrt{\frac{8}{3}}$.
    (Details are omitted due to space constraints.)
    We now derive \eqref{eqn:correlated_gauss_tw_broad}.
    Denote the s.v.d. of $\Sigmamat$ as 
    $
        \Sigmamat = \Umat \Lambdamat \Umat^T
    $,
    where $\Lambdamat\in\Rbb^{p\times p}$ is the diagonal matrix of singular values and
    $\Umat\in\Rbb^{p\times p}$ is unitary.
    Since $\Mmat$ has i.i.d. rows with  $\Nc(0,\Sigmamat)$, it can be decomposed as 
    $
        \Mmat = \Zmat\Lambdamat^{1/2}\Umat^T,
    $
    where entries of $\Zmat\in\Rbb^{n\times p}$ are i.i.d. Gaussian with  $\Nc(0,1)$. 
    Using Property~\ref{prop:eigen_ineq_symm} from Section~\ref{sec:preliminaries}, we obtain
    \begin{equation}\label{eqn:MSigmamatZZT}
        \sigmamin^2(\Mmat) \geq \sigmamin(\Sigmamat)\lambdamin(\Zmat\Zmat\T).
    \end{equation}
    Note that if $\Zmat$ is broad, i.e., $n<p$, then
    $
        \lambdamin(\Zmat\Zmat\T) = \sigmamin^2(\Zmat)
        = \sigmamin^2(\Zmat\T).
    $
    Applying Lemma~\ref{lemma:tracy_widoms_wisdom} to $\Zmat$, we obtain
    \begin{equation}\label{eqn:ZZTpnq}
        \lambdamin(\Zmat\Zmat\T) \geq (\sqrt p -\sqrt n - \Bar{q})_+^2,
    \end{equation}
    with probability at least $1-2e^{-\Bar{q}^2/2}$, $\Bar{q}\geq 0$.
    Note that the bound holds even if $n\geq p$, but it is then informative.
    Using \eqref{eqn:MSigmamatZZT} and \eqref{eqn:ZZTpnq}, we obtain the desired result in \eqref{eqn:correlated_gauss_tw_broad}.

\kern-1em
\subsection{Proof of Theorem~\ref{thm:subg_1}}
\label{proof:thm:sugb_1}
\kern-0.25em
    To find the bound on $\|\Bmat\|$ in Theorem \ref{thm:subg_1},
    we combine \eqref{eqn:B_Bound}, Lemma~\ref{lemma:Abar_A_bound}, Lemma~\ref{lemma:subg_tracy_widom} and the following result:
    \begin{lemma}\label{lemma:subgaussian:minimumsingularvalue:broad}
        Let $\Mmat\in\Rbb^{n\times p}$ be generated as 
        $\Mmat=\Zmat\Lambdamat^{1/2}\Umat\T$, 
        where $\Zmat\in\Rbb^{n\times p}$ has entries i.i.d. with $\Sc(1)$,
        $\Lambdamat\in\Rbb^{p\times p}$ is diagonal with nonnegative entries and $\Umat\in\Rbb^{p\times p}$ is unitary.
        Let $\Sigmamat\in\Rbb^{p\times p}$ denote the corresponding covariance matrix of the rows in $\Mmat$.
        Under $n<p$, the following bound holds with probability at least $1-2e^{-\Bar{q}^2}$, $\Bar{q}\geq 0$:
        \begin{equation}\label{eqn:subg_tw_broad_lemma}
            \sigmamin^2(\Mmat) \geq \sigmamin(\Sigmamat)
            (\sqrt p - CL^2(\sqrt n + \Bar{q}))_+^2,
        \end{equation}
        with $C$ and $L$ as in Lemma~\ref{lemma:subg_tracy_widom}.
    \end{lemma}
    
    \noindent Proof:  By Property \ref{prop:eigen_ineq_symm} of Section \ref{sec:preliminaries}, we have
    \begin{equation} \label{eqn:subgaussian:sigmamin}
        \sigmamin^2(\Mmat) \geq \sigmamin(\Sigmamat)\lambdamin(\Zmat\Zmat\T).
    \end{equation}
    Note that $\sigmamin^2(\Zmat\T)=\lambdamin(\Zmat\Zmat\T)$ for $n<p$. 
    Using \eqref{eqn:subgaussian:sigmamin} with \cite[Thm.~4.6.1]{vershynin2018high}, we obtain the desired result in \eqref{eqn:subg_tw_broad_lemma}.  \qed

    We now continue with the proof of Theorem~\ref{thm:subg_1}. 
    Since $\Amat_k=\Zmat_k\Lambdamat_k^{1/2}\Umat_k$, the rows of $\Amat_k$ are zero-mean sub-gaussian random vectors with the covariance matrix $\Sigmamat_k = \Umat_k\Lambdamat_k\Umat_k\T$ (see Definition~\ref{def:subg_rvar} and \ref{def:subg_rvec}).
    Hence, \eqref{eqn:subg_tracy_widom_tall} holds with $\Sigmamat=\Sigmamat_k$,  $\Mmat = \Amat_k$, 
    $
        \ell(q) =  
        \ell_k(q_k) = CL_k^2\left(\sqrt{\frac{p_k+q_k}{n}} + \frac{p_k+q_k}{n}\right)n\sigmamax(\Sigmamat_k),
    $
    with probability at least $1-2e^{-q_k}$  
    and $L_k\geq 1$ such that
    $
        \psi_{\avec_{i,k}}(\Sigmamat_k,\hvec) \leq 
        L_k \sqrt{\hvec\T \Sigmamat_k\hvec},\text{ any } \hvec\in\Rbb^{p_k\times 1},
    $
    where $\avec_{i,k}\sim\Sc(\Sigmamat_k)$.
    By \eqref{eqn:subg_tw_broad_lemma}, we also have
    $
        \sigmamin^2(\Amat_k)\geq \sigmamin(\Sigmamat_k)(\sqrt{p_k} - CL_k^2(\sqrt{n} + \Bar{q}_k))_+^2,
    $
    with probability at least $1-2e^{-\Bar{q}_k^2}$, if $n<p_k$.
    To find the bound on $\|\Bmat\|$ in \eqref{eqn:B:thm:subg_1}, we
    plug in the upper bound on $\sigmamax^2(\Amat_k)$ for each $\Amat_k$,
    and the respective lower bound for $\sigmamin^2(\Amat_k)$, depending on whether $\Amat_k$ is broad or tall.
    Using  $\Kc=\{k:n<p_k\}$,
    and Property \ref{sec:prob_intersect} of Section \ref{sec:preliminaries},
    we find the desired probability bound     $
        1 - \sum_{k=1}^K 2e^{-q_k} - \sum_{k\in\Kc}2e^{-\Bar{q}_k^2}.
    $

\kern-1em
\subsection{Proof of Theorem \ref{thm:subg_2}}
\label{proof:thm:subg_2}
\kern-0.25em
    The proof follows a similar line of argument with the proof of Theorem~\ref{thm:subg_1}. 
    In particular, we use \eqref{eqn:subg_tracy_widom_tall} with $\Sigmamat=\Sigmamat_k$, $\Mmat = \Amat_k$. %
    The probability expression is found using the probability bound for \eqref{eqn:subg_tracy_widom_tall} for each $k$, 
    i.e., $1-2e^{-q_k}$,  and Property~\ref{sec:prob_intersect} of Section \ref{sec:preliminaries}. 
    We omit the details due to space constraints.

%% file: appendix/preliminaries.tex
\subsection{Preliminaries}\label{sec:preliminaries}
\kern-0.25em
This section provides a collection of properties that are used frequently in different proofs: 

\begin{enumerate}[wide, labelindent=0pt, label=(\alph*)]
    \item 
        \label{sec:prob_intersect}
        Given $c_k$, $\krange$,
        with  $\Pr(c_k)\geq 1-\rho_k$, the probability of intersection can be bounded as
        \begin{equation}\label{eqn:probability_prop}
            \Pr\left({\bigcap_{k=1}^K} c_k \right) 
            \geq 1 - { \sum_{k=1}^K} \rho_k.
        \end{equation}
    \item 
        \label{hölders_trick}
        Partition a symmetric matrix $\Mmat \!\succeq\! 0$ as
        $
        \Mmat \!=\! 
            \begin{bmatrix}
                \Mmat_{11}      & \Mmat_{12} \\ 
                \Mmat_{12}\T    & \Mmat_{22}
            \end{bmatrix}.
        $
        Then,
            $\|\Mmat\| \leq \|\Mmat_{11}\| + \|\Mmat_{22}\|.$
    \item 
        \label{sec:prop_spectral_norm_bounds}
        Let $\Amat$, $\Bmat$ be two real square matrices. If $\|\Amat-\Bmat\| \leq q$,  then $\sigmamin(\Amat) \geq \sigmamin(\Bmat) - q$.
    \item 
        \label{prop:eigen_ineq_symm}
        Let $\Lambdamat\in\Rbb^{p\times p}$ be diagonal with the entries
        $\mu_p\geq \cdots \geq \mu_1 \geq 0$,
        and let $\Umat\in\Rbb^{p\times p}$ be unitary.
        Let $\Amat\in\Rbb^{n\times p}$ have the following decomposition in terms of another matrix $\Zmat\in\Rbb^{n\times p}$ as $
            \Amat = \Zmat\Lambdamat^{1/2}\Umat^T.
        $
        Then 
        $
            \sigmamin^2(\Amat) \geq \mu_1 \lambda_{\min}(\Zmat\Zmat\T).
        $
    \item \label{prop:gauss_identity}
    Let $\Mmat\in\Rbb^{n\times p}$ be generated as 
    $\Mmat=\Zmat \Lambdamat^{1/2} \Umat\T$,
    where $\Zmat\in\Rbb^{n\times p}$ has i.i.d. entries with $\Nc(0,1)$,
    $\Lambdamat\in\Rbb^{p\times p}$ is diagonal and positive definite,
    and $\Umat\in\Rbb^{p\times p}$ is unitary.
    Then, $\Mmat\p\Mmat=\eye{p}$ and $\Mmat\Mmat\p=\eye{n}$ with probability (w.p.) one, if $n\geq p$ or $p\geq n$, respectively.
\end{enumerate}

We now present the proofs for these properties. 

\subsubsection{Proof of Property~\ref{sec:prob_intersect}}
 We observe that
    \begin{equation}\label{eqn:probability_prop_proof}
        \Pr\left({ \bigcap_{k=1}^K} c_k\right) 
        = 1 - \Pr\left(\left({ \bigcap_{k=1}^K} c_k\right)^c\right) 
        = 1 - \Pr\left({ \bigcup_{k=1}^K} c_k^c\right)
    \end{equation}
    Using the union bound we have that
   $
        \Pr\left(\bigcup_{k=1}^K c_k^c\right) \leq \sum_{k=1}^K \Pr(c_k^c).
    $
    By definition, $\Pr(c_k) = 1-\Pr(c_k^c) \geq 1 - \rho_k$,
    or equivalently, $\Pr(c_k^c) \leq \rho_k$. Hence,
    $
        \Pr\left(\bigcup_{k=1}^K c_k^c\right) \leq \sum_{k=1}^K \rho_k.
    $
    Using this inequality together with \eqref{eqn:probability_prop_proof} yields to the desired inequality after re-arranging the terms. 
    
\subsubsection{Proof of Property~\ref{hölders_trick}} 
    See \cite[Proposition~8.3]{HalkoMartinssonTropp_2011}. 

\subsubsection{Proof of Property~\ref{sec:prop_spectral_norm_bounds}}
    Let $\uvec$ be any vector such that $\|\uvec\|=1$. 
    Combining 
    $\|\Amat - \Bmat\| = 
    \|\Amat-\Bmat\|\|\uvec\| 
    \geq 
    \|\Amat\uvec - \Bmat\uvec\|$ 
    and 
    $ \|\Amat-\Bmat\| \leq q$, we obtain   $ q \geq \|\Amat\uvec - \Bmat\uvec\|$. Applying reverse triangle inequality, we have  $ q  \geq |\|\Amat\uvec\|-\|\Bmat\uvec\|| $ which yields to  
    $
        q \geq  \|\Amat\uvec\|-\|\Bmat\uvec\|  \geq - q.
    $
    Rearranging the right-hand side inequality, we obtain
    $
        \|\Amat\uvec\| \geq \|\Bmat\uvec\| - q.
    $
    Let $(\lambda_{\min}(\Amat\T\Amat),\bar{\uvec})$ with $\|\bar{\uvec}\|=1$ be the eigenpair corresponding to the smallest eigenvalue  of $\Amat\T\Amat$. Then,
    \begin{align}
        \|\Amat \bar{\uvec}\| &= \sqrt{\bar{\uvec}\T\Amat\T\Amat\bar{\uvec}} = \sqrt{\lambda_{\min}(\Amat\T\Amat)} \\
        &\geq \sqrt{\bar{\uvec}\T\Bmat\T\Bmat\bar{\uvec}} - q \geq \sqrt{\lambda_{\min}(\Bmat\T\Bmat)} - q.
    \end{align}
    Note that 
    $\sqrt{\lambda_{\min}(\Mmat\T\Mmat)} = \sigmamin(\Mmat)$
    for any $\Mmat\in\Rbb^{n\times n}$, so
    $
        \sigmamin(\Amat) \geq \sigmamin(\Bmat) - q 
    $.

 \subsubsection{Proof of     Property~\ref{prop:eigen_ineq_symm}}   
    Writing $\Amat\Amat\T$ in terms of $\Zmat$
    \begin{equation}
        \Amat\Amat\T = \Zmat\Lambdamat^{1/2}\Umat\T\Umat\Lambdamat^{1/2}\Zmat\T
        = \Zmat \Lambdamat \Zmat\T,
    \end{equation}
    together with adding and subtracting $\mu_1\Zmat\Zmat\T$, we obtain
    \begin{equation}\label{eqn:prel:AAT}
        \Amat\Amat\T = \mu_1\Zmat\Zmat\T + \Zmat(\Lambdamat - \mu_1\eye{p})\Zmat\T.
    \end{equation}
    Let $(\lambda_{\min}(\Amat\Amat\T), \vvec)$ be the eigenpair corresponding to the smallest eigenvalue of $\Amat\Amat\T$. We  now evaluate $\vvec\T\Amat\Amat\T\vvec$ using \eqref{eqn:prel:AAT} to obtain
    \begin{align} 
        \lambda_{\min}(\Amat\Amat\T) &= \mu_1 \vvec\T \Zmat\Zmat\T\vvec + \vvec\T\Zmat(\Lambdamat - \mu_1\eye{p})\Zmat\T\vvec, \\
        & \geq \mu_1\lambda_{\min}(\Zmat\Zmat\T),
    \end{align}
    where we have used that  $\vvec\T\Zmat(\Lambdamat - \mu_1\eye{p})\Zmat\T\vvec \geq 0$.
    Note that $\sigmamin^2(\Amat) \geq \sigmamin(\Amat\Amat\T) = \lambdamin(\Amat\Amat\T)$, and we have
    $
        \sigmamin^2(\Amat) \geq \mu_1 \lambdamin(\Zmat\Zmat\T).
    $
    
\subsubsection{Proof of Property \ref{prop:gauss_identity}}
    By \cite[eqn (3.2)]{rudelson_non-asymptotic_2010},  $\Zmat$ is full rank w.p. 1.
    Hence, $\Mmat$ is full rank since it is the product of full rank matrices.
    Thus, with $n\geq p$,  $\Mmat\T\Mmat\in\Rbb^{p\times p}$ is full rank,
    i.e. invertible.
    Hence $\Mmat\p\Mmat = (\Mmat\T\Mmat)\p\Mmat\T\Mmat=\eye{p}$.
    A similar line of argument holds for $p\geq n$ with $\Mmat\Mmat\p=\eye{n}$.
    

%% file: props/lemma_Abar_A_bound.tex
\begin{lemma}\label{lemma:Abar_A_bound}
For any matrix $\Amat=[\Amat_1, \cdots, \Amat_K]\in\Rbb^{n\times p}$
and $\Abar = [\Amat_1\p; \cdots; \Amat_K\p]\in\Rbb^{p\times n}$,
the following bound holds:
\begin{align}
    \|\Abar \Amat\|^2
    &\leq K \! 
    +\!
    \sum_{k=1}^K
    \sum_{\substack{i=1\\i\neq k}}^K
    \frac{\sigma_{\max}^2(\Amat_k)}{\sigma_{\min+}^2(\Amat_i)}, 
    \label{eqn:Abar_A_bound_upper}
\end{align}
where $\sigma_{\min+}(\Amat_i)$ denotes the smallest non-zero singular value of $\Amat_i$.
\end{lemma}